\documentclass[10pt,twocolumn,letterpaper]{article}

\usepackage{cvpr}
\usepackage{times}
\usepackage{epsfig}
\usepackage{graphicx}
\usepackage{amsmath}
\usepackage{amssymb}
\DeclareMathOperator*{\argmin}{arg\,min}
\usepackage{booktabs,makecell}
\usepackage{multirow}

\usepackage{xcolor}

\newcommand{\ourmethod}{Surface HOF}
\usepackage[pagebackref=true,breaklinks=true,letterpaper=true,colorlinks,bookmarks=false]{hyperref}

\cvprfinalcopy 

\def\cvprPaperID{8414} 

\ifcvprfinal\fi

\title{Surface HOF: Surface Reconstruction from a Single Image Using Higher Order Function Networks}
\author{
Ziyun Wang, Volkan Isler, Daniel D. Lee\\
Samsung AI Center - New York\\
{\tt\small saic-ny@samsung.com}
}
\begin{document}

\maketitle



\begin{abstract}
We address the problem of generating  a high-resolution surface reconstruction from a single image. Our approach is to learn a Higher Order Function (HOF) which takes an image of an object as input and generates a mapping function. The mapping function takes samples from a canonical domain (e.g. the unit sphere) and maps each sample to a local tangent plane on the 3D reconstruction of the object. Each tangent plane is represented as an origin point and a normal vector at that point. By efficiently learning a continuous mapping function, the surface can be generated at arbitrary resolution in contrast to other methods which generate fixed resolution outputs. We present the {\bf Surface HOF} in which  both the higher order function and the mapping function are represented as neural networks, and train the networks to generate reconstructions of PointNet objects. Experiments show that Surface HOF is more accurate and uses more efficient representations than other state of the art methods for surface reconstruction. 
Surface HOF is also easier to train: it requires minimal input pre-processing and output post-processing and generates surface representations that are more parameter efficient. Its accuracy and convenience make Surface HOF an appealing method for single image reconstruction.
\end{abstract}

\section{Introduction}

\begin{figure}[h!]
\begin{center}
    \centering
    \includegraphics[width=0.45\textwidth]{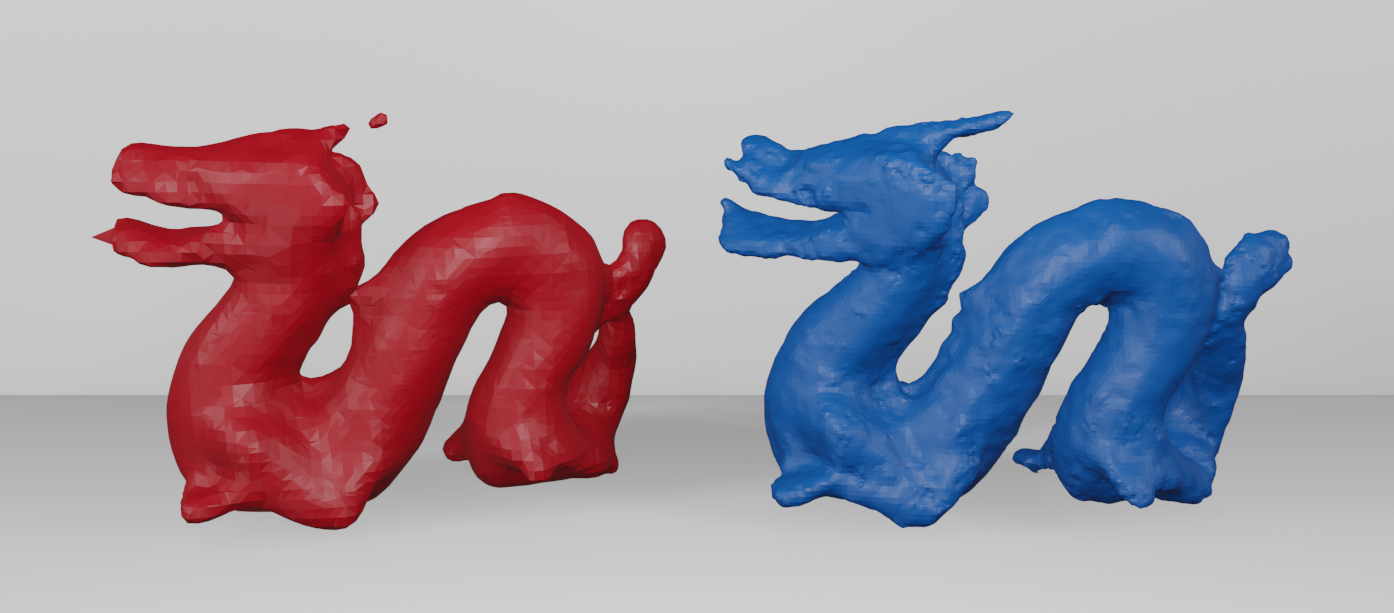}
    \caption {
    To demonstrate the continuous nature of Surface HOF,  we trained the network on the dragon model using a single mapping function.
    The red dragon on the left is obtained by mapping 1000 samples from the unit sphere. The blue dragon on the right is obtained by mapping 10 thousand samples using the same mapping function. 
} 
    \label{fig:two_dragons}
\end{center}%
\end{figure}

Deep learning methods have now been established as state of the art for solving fundamental computer vision problems such as object detection and image segmentation. Recently, there has been significant research activity around extending their successes to processing and generating three-dimensional (3D) data. Some representative tasks in this domain include reconstructing the 3D model of an object from one or more images~\cite{tatarchenko2019single}, completing a partial 3D model~\cite{qi2017pointnet} or decomposing a 3D model into its parts~\cite{mo2019partnet}. 




There are many applications  which require continuous surface representations of objects.  For example, high resolution surface normals provide key information  to compute appearance information for graphics rendering. Similarly, in robotic manipulation, normals are needed to compute the effect of frictional forces. Motivated by these applications, an active research topic at present is to design architectures which can generate continuous surface representations. Recent successes in this domain include
AtlasNet~\cite{groueix2018atlasnet} which represents the object as a union of manifolds, 
Pixel2Mesh~\cite{wang2018pixel2mesh} which generates a mesh using graph convolutions over a fixed input mesh and DeepSDF~\cite{park2019deepsdf} which outputs a signed distance function representation of the object.

\begin{figure*}[t]
\begin{center}
   \includegraphics[width=0.9\textwidth]{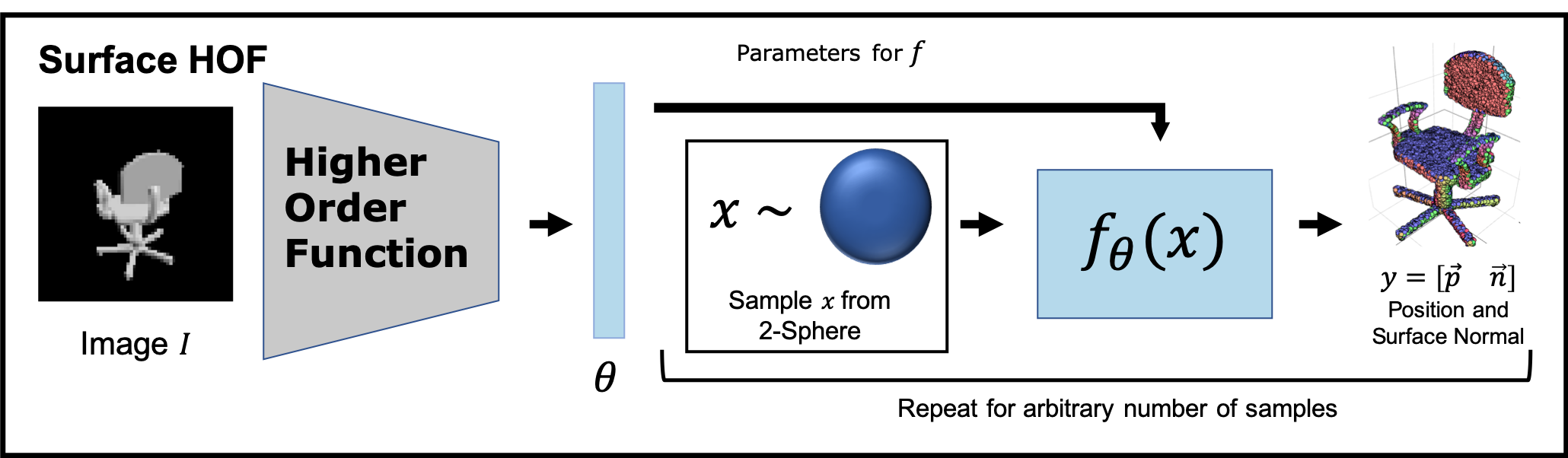} 
   \includegraphics[width=0.9\textwidth]{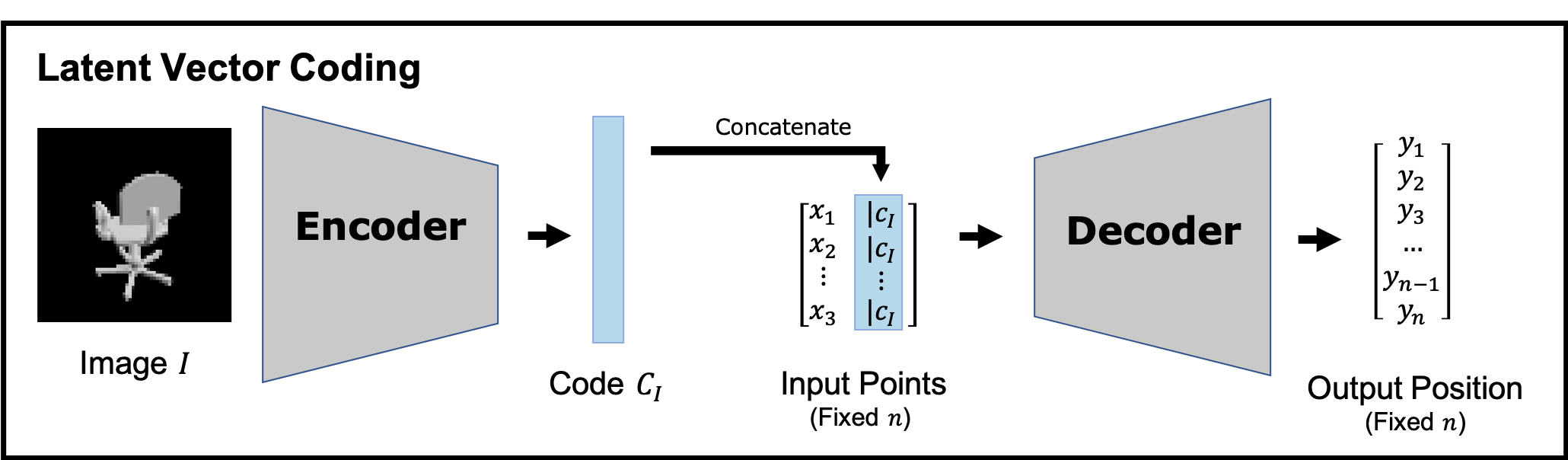}
\end{center}
   \caption{Top: Surface HOF (our approach). The Higher Order Function generates the parameters for $f_\theta$ which can be used to generate output $y = f_\theta(x)$ for any $x$. Bottom: Latent Vector Encoding. The encoder generates a codeword $c_I$ for image $I$. The same codeword is appended to a fixed number of $n$ parameters ($x_i$) to estimate $y_i$ corresponding to $x_i$, $i=1, \ldots,n$.
    \label{fig:our_approach}}
\end{figure*}

In this paper, we build on this body of work and present a new approach for surface reconstruction based on higher order function representations. Our primary focus is on 3D model generation from a single image. Our approach, summarized in Figure~\ref{fig:our_approach}, learns a mapping function $f_\theta$ which maps each sample from a canonical domain to  the tangent plane on a particular location of the object. The function $f_\theta$ itself is generated by a higher order function $g$ which takes the input image and outputs $\theta$ where $\theta$ are function parameters of $f_\theta$ for the specific input image.  Both $f_\theta$ and $g$ are represented as neural networks.  Hence, $\theta$ corresponds to the weights of the neural network representing $f_\theta$.

Figure~\ref{fig:two_dragons} illustrates the continuous nature of Surface HOF: 
The two dragons in the figure are generated using the same $f_\theta$. The one on the left is the image of 1000 randomly sampled points on the unit sphere, whereas the one the right is obtained from the image of 10,000 points mapped by the \emph{same} function. The network outputs a tangent plane for each point which is converted to a mesh for rendering purposes. 

In order to establish Surface HOF as an effective model for surface reconstructon, we compare it with state of the art approaches which generate surfaces from a single image. Surface HOF outperforms them in terms of point quality and the quality of normals.
Our results indicate that Surface HOF has many appealing properties: Since we learn a local mapping function with a global objective function, it is capable of generating accurate surfaces of complex objects. Data generation is fairly straightforward as the ground truth requires only point samples and normal vectors which can be obtained from local patches like triangles. The network requires significantly smaller number of parameters compared to state of the art approaches. Therefore training is also easier. 


\section{Related Work}

\noindent
In this section, we present an overview of related work. 

\paragraph{Shape Encoding} Mapping an object shape to a fixed-sized latent vector is a common approach for encoding the shape information of an object. Point-based encoders \cite{qi2017pointnet, qi2017pointnetpp} extract global features of a point set to a fixed-sized codeword $z$. Eventually, $z$ is decoded into task-specific representations \cite{park2019deepsdf, groueix2018atlasnet, dai2017shape, choy20163d, yang2018foldingnet}. A recent alternative to fixed size encoding is Higher Order Function (HOF) \cite{anon2019} where the extracted shape information is encoded as the weights of another network. In this paper, we extend the HOF paradigm from generating point sets to generating surfaces, and compare Surface HOF to methods that generate surfaces from images.

\paragraph{3D Reconstruction from a Single RGB Image} 3D reconstruction is a well-studied area in multi-view geometry. Structure from Motion (SFM) approaches reconstruct the visible parts of the 3D model of an object by exploiting the geometric consistency between multiple views. Recently, there is a growing interest in inferring the full 3D shape information from a single image. The main challenge in this problem is that only a partial view is observed and that relative depth of the points cannot be recovered from a single image. Recent learning-based approaches exploit the ability of deep neural networks to extract  rich semantic information from single-view images. Multiple representations are proposed to represent the predicted 3D shape:

Voxel-based methods \cite{choy20163d} discretize the object domain into a finite-resolution occupancy grid. These methods take advantage of the fact that Convolutional Neural Networks (CNN) excel at processing data on a regular grid. Despite the ease of training, the computational cost increases cubically with the increase of resolution. Therefore, most voxel-based methods are trained on a low-resolution voxelization of the object.  Methods such as \cite{hane2017hierarchical, tatarchenko2019single, tatarchenko2017octree} address the problem by adopting a coarse-to-fine representation.

A point set is also popular choice of 3D shape representation. Contrary to voxels, point sets can be sampled in arbitrary resolution. Some works attempt to map points from regular domain to the object and produce visually-pleasing point sets~\cite{qi2017pointnet}. While this line of work successfully recovers the general shape of an object, detailed surface properties of the object do not necessarily propagate to the final reconstruction. To address this deficiency, Qi et al.~\cite{qi2017pointnetpp} and He et al.,  \cite{he2019geonet} improved the PointNet encoding by incorporating Euclidean neighborhood and geodesic distance information. However, recovering the full geometric details solely from a point set remains a challenging problem. Zhang et al. \cite{zhang2018genre} propose to directly use a spherical map to represent the object. However, this assumes the existence of a valid spherical projection of the object that covers the entire object surface. 
Recently, \cite{wang2018pixel2mesh,groueix2018atlasnet, yang2018foldingnet, kanazawa2018learning} focus on directly working with pre-defined grid domain and predicting its deformation conditioned on a single image.

Atlas-Net!\cite{groueix2018atlasnet, ben2018multichart} learns an atlas composed of multiple sheets to represent the object surface. While the method has the ability to represent arbitrary genus shapes, it relies on the arbitrary initialization of $K$ patches that cover the entire object. The reconstruction quality is reported to be heavily dependent on the choice of $K$ which can be as large as 125. 

DeepSDF~\cite{park2019deepsdf} and Deep Level Sets \cite{michalkiewicz2019deep} learn continuous functions which implicitly represent the surface of an object. However, such methods require careful pre/post processing of the ground truth for training. For input dense, carefully selected samples near the surface with correct signs must be generated.
Once the network is trained, in order to extract the normals, post-processing operations such as voxelization and marching cubes need to be performed. 
Pixel2Mesh \cite{wang2018pixel2mesh} uses a Graph Neural Network to move the vertices of a pre-triangulated ellipsoid. Additional constraints, such as surface normal and edge length, are used to improve the overall quality of output mesh. The method, however, is limited to genus-0 objects due to the topology of the mapped ellipsoid. GEOMetrics~\cite{smith2019geometrics} introduces face splitting operations to refine the mesh. Mesh RCNN~\cite{gkioxari2019mesh} also uses graph convolutions and combines segmentation with surface estimation. 

After presenting the details of Surface HOF in the next section, we compare it with state of the art methods which are able to generate a surface from a single RGB image. With a very simple decoder (3 layer perceptron with 256 parameters at each layer), Surface HOF achieves better point cloud and surface normal accuracy compared to much more sophisticated state of the art methods.










\section{Method}

In this section, we present an overview of the Surface HOF method for generating a 3D model from a single image. 
The general HOF architecture can be viewed as a generalization of Latent Vector Code based architectures composed of an encoder and a decoder (Figure~\ref{fig:our_approach}-bottom). In this approach, the encoder generates a latent codeword $c_I$ for an input image $I$. This code is usually concatenated with each element $x_i$ of a fixed number of input parameters $x_i$, $i=1, \ldots, n$. For example, $x_i$ can be the coordinates where a shape related value such as occupancy will be estimated. 
The decoder takes a stack of $n$ vectors $[x_i | c_I]$ and estimates shape information at each $x_i$. More recent methods inject the codeword at multiple layers across the decoder. Regardless, the codeword $c_I$ is the same across all $x_i$. In this setting, the codeword $c_I$ acts as a per-image bias of the input to the decoder whose weights remain fixed after training.

HOF is a generalization of this paradigm because rather than a fixed bias, the higher order function outputs weights of the decoder (Figure~\ref{fig:our_approach}-top). In contrast to the LVC approach, the weights of the mapping function are estimated for each input image. The mapping function $f_\theta$ is trained using random samples of input and target domains at each iteration. The output of $f_\theta$ is a point vector and a direction which is local. The network  is trained with a global loss function. Therefore, the representation resolution of Surface HOF is only limited by the resolution of the training set. It can learn to output surfaces for complex shapes. 


Specific details about the implementation and training of the Surface HOF architecture are presented next.

\subsection{Architecture}
To achieve the task of single image reconstruction, we use a CNN for encoding the semantic and geometric information of an image. We implement a DenseNet-based convolutional neural network introduced in \cite{huang2017densely}. We use three convolution layers, each followed by a densely connected convolution layer. The output of the CNN is followed by a layer with 1024 hidden units. The mapping function  $f_\theta$ is composed of 3 fully connected layers, each with 128 hidden units. Rectified Linear Units (ReLU) are used as the activation functions of all layers. 

In our implementation, the input point is mapped uniformly from the surface of a 3-dimensional sphere (represented as a three dimensional vector) and mapped to a 6-dimensional output $[p \; v]$ representing the tangent plane where $p$ is position vector and $v$ is a direction vector. 

\subsection{Losses}
\paragraph{Chamfer Loss} Chamfer distance is a commonly used loss function for evaluating the distances between two unordered sets of points. By minimizing the cross-set distances of the generated point set and ground truth point set, the network learns to map points closer to the surface of the desired surface. Here, we use symmetric Chamfer distance as our loss function. Symmetric Chamfer distance is defined as:
\begin{align}
\mathcal{L}_{CD}(X, Y) & =   \mathcal{L}_{XY} + \mathcal{L}_{YX} \\
 & = \frac{1}{|X|}\sum_{x\in X}{\min_{y\in Y}{||x - y||_2}} \\ &+ \frac{1}{|Y|}\sum_{y\in Y}{\min_{x\in X}{||x - y||_2}}
\end{align}

Note that here we use the norm distance rather than the squared distance between two points in Chamfer distance, which is different from the evaluation metric in Equation~\ref{eq:1}.

\paragraph{Cosine Surface Loss} Cosine distance is used to measure the angle between the predicted surface normal and the ground truth surface normal. The cosine distance loss is defined as:
\begin{align}
\mathcal{L}_{cos}(X_{gt}, X_{pred}) &= 1 - \frac{1}{|X_{gt}|}\sum_{x\in X_{gt}}{|{\Vec{n_x}\cdot{\Vec{n}_{\theta(x, X_{pred})}}}|} \\
\theta(x, Y) &= \argmin_{y \in Y} ||x - y||_2^2 
\end{align}
In training, we find that having two way cosine distance losses affects the training performance negatively. Having all of the ground truth surface normal contribute to the loss function introduces significant noise into the supervision signal. Therefore, we use only one-way cosine surface loss. The final loss function is computed as the weighted sum of the Chamfer loss and the 
Cosine Surface Loss:
\begin{align}
\mathcal{L} = \lambda_{CD} * \mathcal{L}_{CD} + \lambda_{cos} * \mathcal{L}_{cos}
\end{align}

\subsection{Training} 

In practice, we find that the input dimension of the images has limited effects on the reconstruction quality. Therefore, we linearly interpolate the images to size $64 \times 64$ to speed up our training. We train 20 epochs using Adam optimizer on a single NVidia GTX 1080ti GPU. The learning rate is $1e-5$ and the batch size is 1. Since the quality of surface normal estimation relies on the convergence of the Chamfer distance, we use the loss weight combination of $\lambda_{CD} = 1$ and $\lambda_{cos} = 0.1$. In each iteration, we  sample 1000 points uniformly at random from the surface of a sphere and map these points to the target object. By repeatedly sampling from the sphere, we approximate a continuous mapping from the domain of the spherical surface to the object surface.

\section{Experiments} 

\begin{table}
\centering
\begin{tabular}{c c c c}
\hline
Mesh R-CNN\cite{gkioxari2019mesh} & P2M\cite{wang2018pixel2mesh} & P2M Code & \ourmethod{} \\
\hline
 0.284 & 0.591 & 0.482 & \textbf{0.237}\\
\hline
\end{tabular}
\caption{Mean Chamfer distance comparison. All values are multiplied by 1000 for ease of comparison.\label{tab:meancompare}}
\end{table}

\begin{table*}[t]
\centering
\begin{tabular}{l c c c c c c}
\hline
Category & 3D-R2N2\cite{choy2016r2n2}  & PSG\cite{fan2017point}  & N3MR\cite{kato2018neural} & P2M\cite{wang2018pixel2mesh} & P2M Code & \ourmethod{}  \\
\hline \hline
plane       & 0.895 & 0.430 & 0.450 & 0.477 & 0.372 & \textbf{0.109}\\
bench       & 1.891 & 0.629 & 2.268 & 0.624 & 0.473 & \textbf{0.211}\\
cabinet     & 0.735 & 0.439 & 2.555 & 0.381 & 0.331 & \textbf{0.225}\\
car         & 0.845 & 0.333 & 2.298 & 0.268 & 0.242 & \textbf{0.158}\\
chair       & 1.432 & 0.645 & 2.084 & 0.610 & 0.507 & \textbf{0.282}\\
monitor     & 1.707 & 0.722 & 3.111 & 0.755 & 0.569 & \textbf{0.291}\\
lamp        & 4.009 & 1.193 & 3.013 & 1.295 & 1.032 & \textbf{0.387}\\
speaker     & 1.507 & 0.756 & 3.343 & 0.739 & 0.633 & \textbf{0.403}\\
firearm     & 0.993 & 0.423 & 2.641 & 0.453 & 0.382 & \textbf{0.118}\\
couch       & 1.135 & 0.549 & 3.512 & 0.490 & 0.441 & \textbf{0.246}\\
table       & 1.116 & 0.517 & 2.383 & 0.498 & 0.385 & \textbf{0.232}\\
cellphone   & 1.137 & 0.438 & 4.366 & 0.421 & 0.342 & \textbf{0.214}\\
watercraft  & 1.215 & 0.633 & 2.154 & 0.670 & 0.558 & \textbf{0.204}\\
\hline
mean  & 1.445& 0.593 & 2.629 & 0.591 & 0.482 & \textbf{0.237}\\
\hline
\hline
\end{tabular}
\caption{Symmetric Chamfer distance evaluated on the ShapeNet Test set. All numbers are multiplied by 1000 for ease of comparison. Lower is better. \label{tab:chamfer_p2m}
 }
\end{table*}

\begin{table*}[t]
\centering
\begin{tabular}{l || c c|| c c}
\hline
& \multicolumn{2}{c}{$\tau$} & \multicolumn{2}{c}{$2\tau$} \\

Category & P2M\cite{wang2018pixel2mesh} & \ourmethod{} & P2M\cite{wang2018pixel2mesh} & \ourmethod{}  \\
\hline \hline
plane       & 71.12 & \textbf{81.77} & 81.38 & \textbf{90.47}\\ 
bench       & 55.57 & \textbf{63.41} & 71.86 & \textbf{79.38} \\ 
cabinet     & \textbf{60.39} & 59.64 & 77.19 & \textbf{79.58}\\ 
car         & \textbf{67.86} & 66.02 & 84.15 & \textbf{84.86}\\ 
chair       & \textbf{54.38} & 53.60 & 70.42 & \textbf{72.29}\\ 
monitor     & 51.31 & \textbf{54.85} & 67.01 & \textbf{73.47}\\ 
lamp        & 48.15 & \textbf{55.16} & 61.50 & \textbf{70.22}\\ 
speaker     & \textbf{48.84} & 47.19 & 65.61 & \textbf{67.72}\\ 
firearm     & 73.20 & \textbf{80.53} & 83.47 & \textbf{89.60}\\ 
couch       & \textbf{51.90} & 51.67 & 69.83 & \textbf{73.64} \\ 
table       & \textbf{66.30} & 65.98 & 79.20 & \textbf{81.36}\\ 
cellphone   & \textbf{70.24} & 68.19 & 82.86 & \textbf{84.23}\\ 
watercraft  & 55.12 & \textbf{64.06} & 69.99 & \textbf{79.98}\\ 
\hline
mean & 59.72 & \textbf{62.70} & 74.19 &  \textbf{78.98}\\
\hline
\hline
\end{tabular}
\caption{F-score evaluated on the ShapeNet Test set. Higher is better. $\tau=10^{-4}$\label{tab:f1}
 }
\end{table*}

\begin{table*}
\begin{tabular}{c c c c}
  \includegraphics[width=0.23\textwidth]{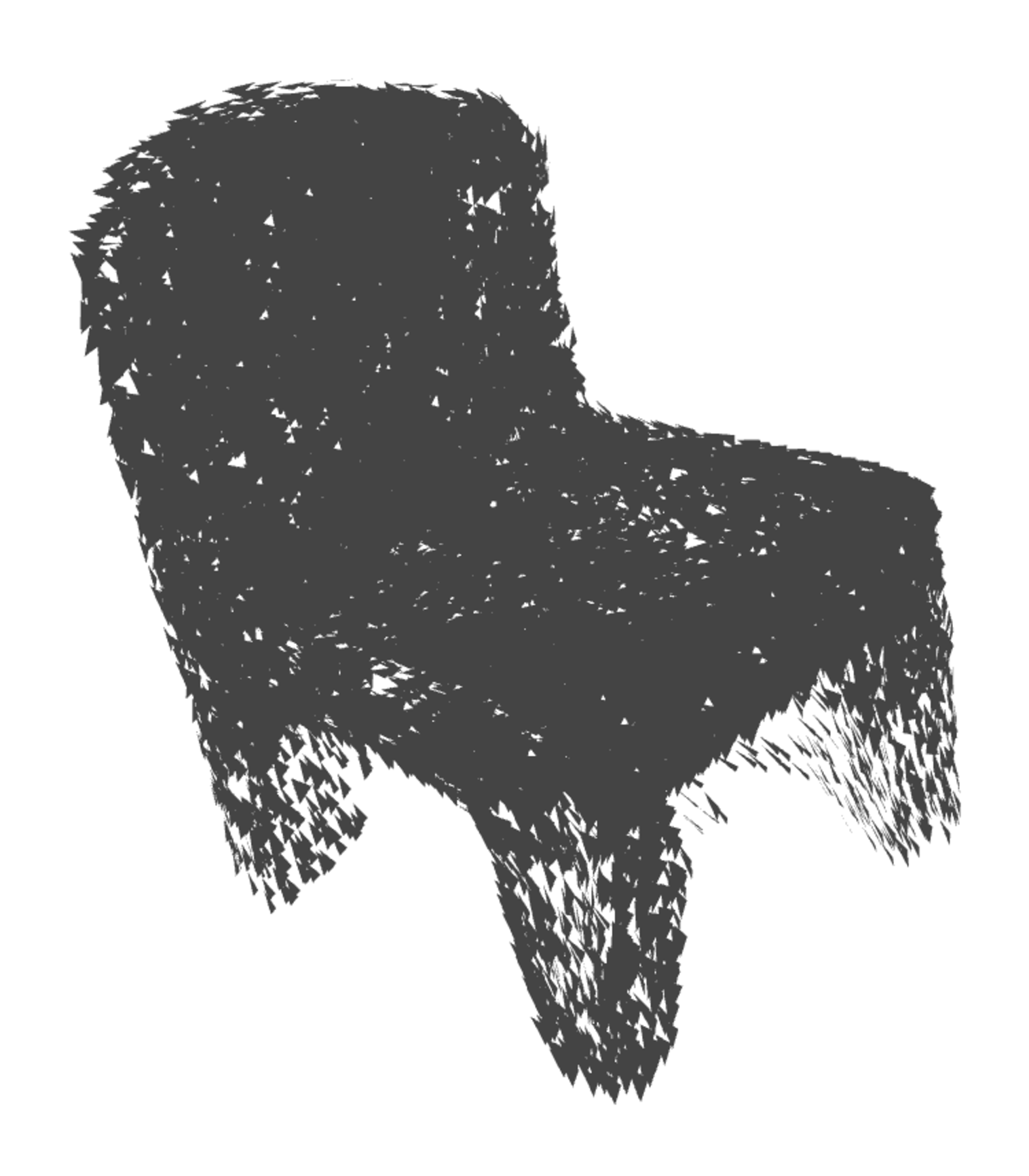} & \includegraphics[width=0.23\textwidth]{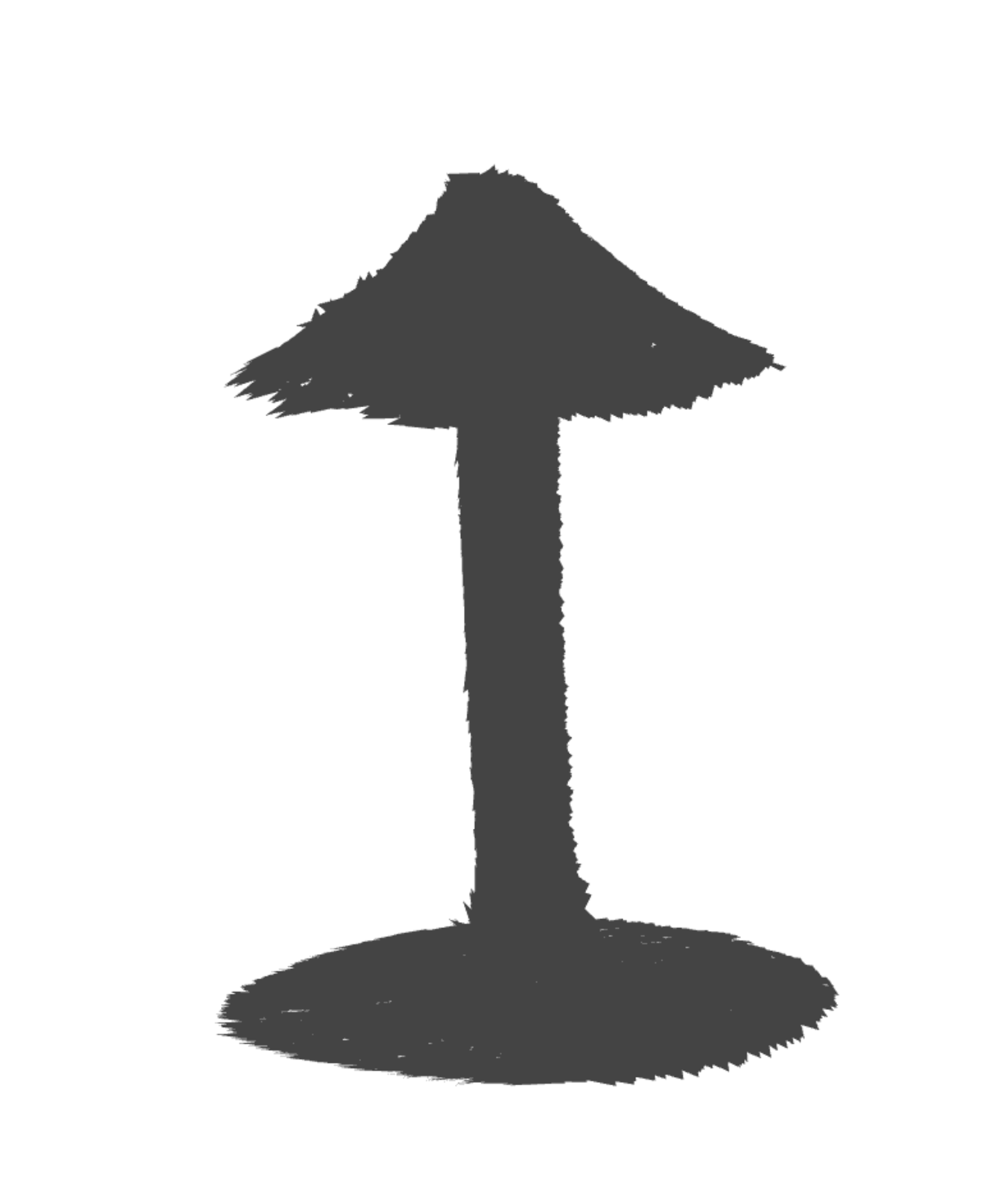} & 
  \includegraphics[width=0.23\textwidth]{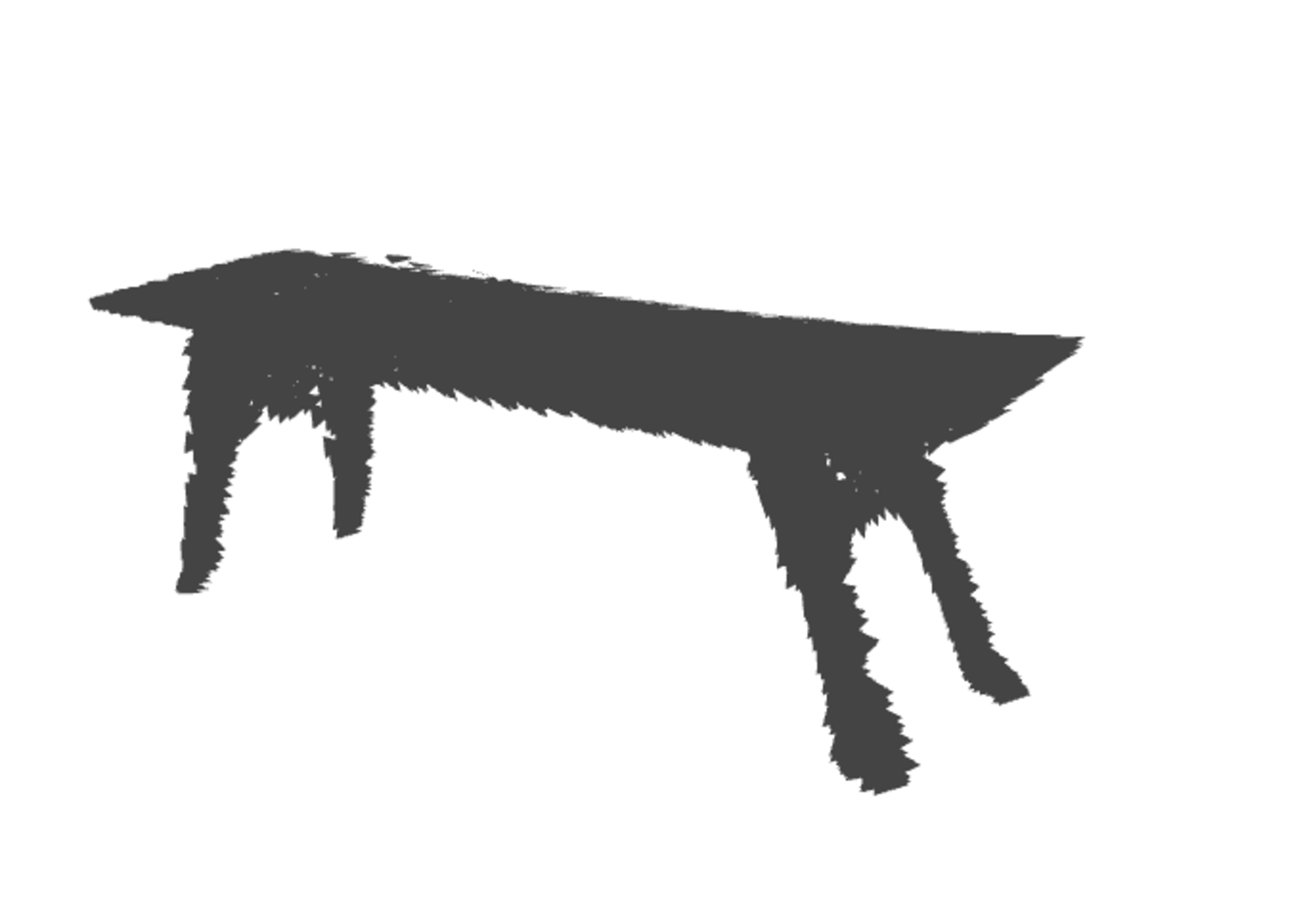} &
  \includegraphics[width=0.23\textwidth]{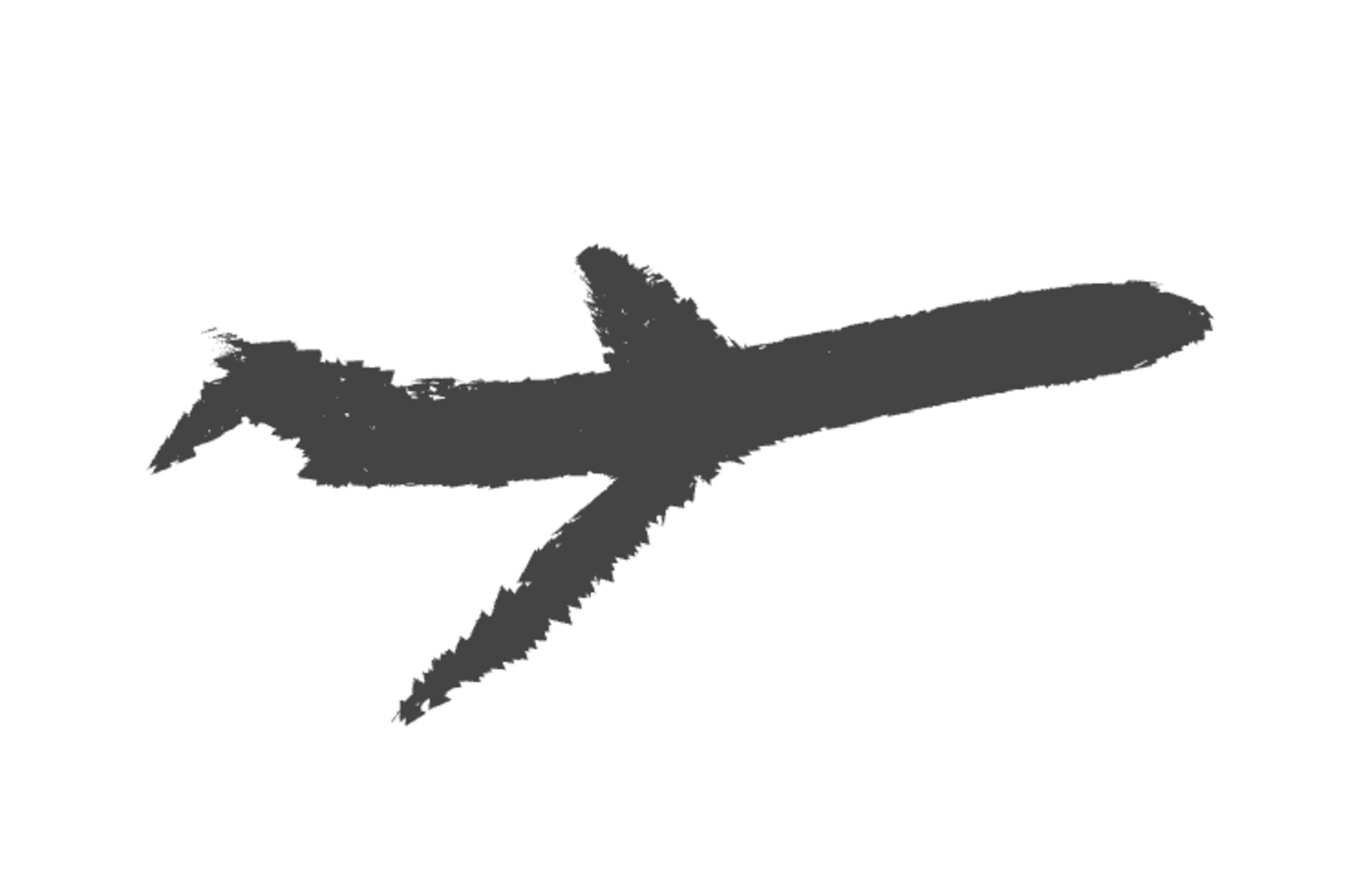}
   \\ 
  
  \includegraphics[width=0.23\textwidth]{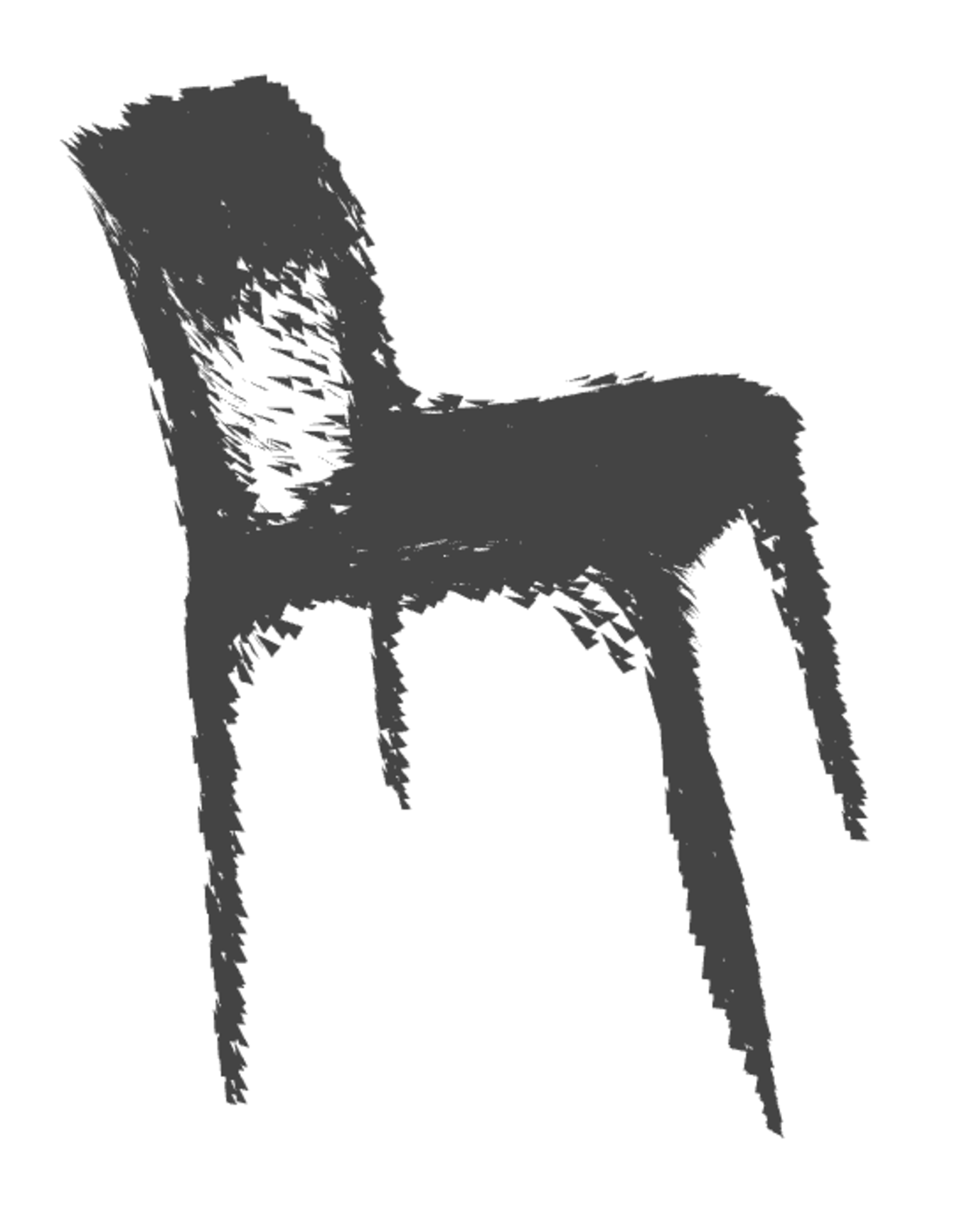} & \includegraphics[width=0.23\textwidth]{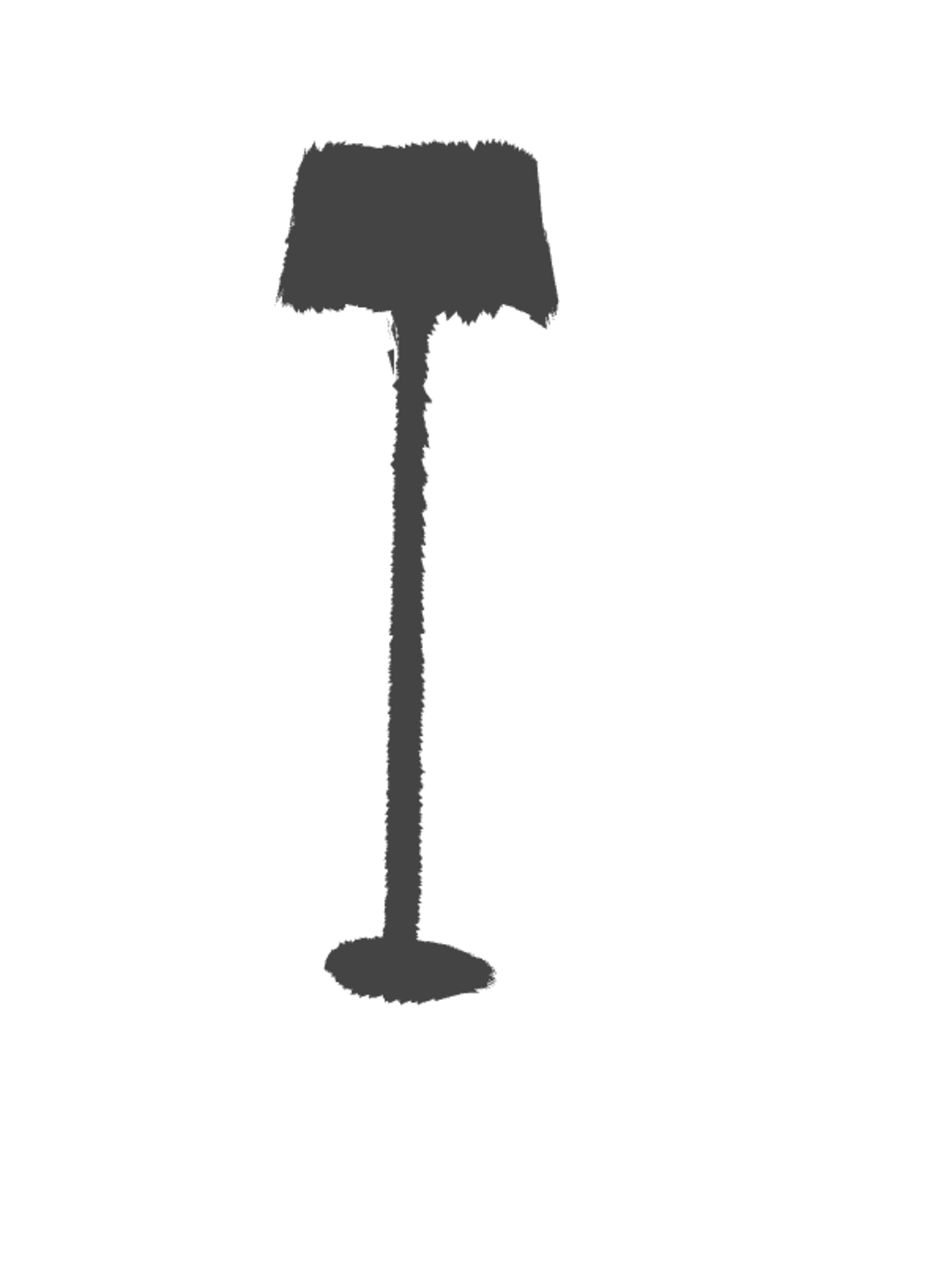} & 
  \includegraphics[width=0.23\textwidth]{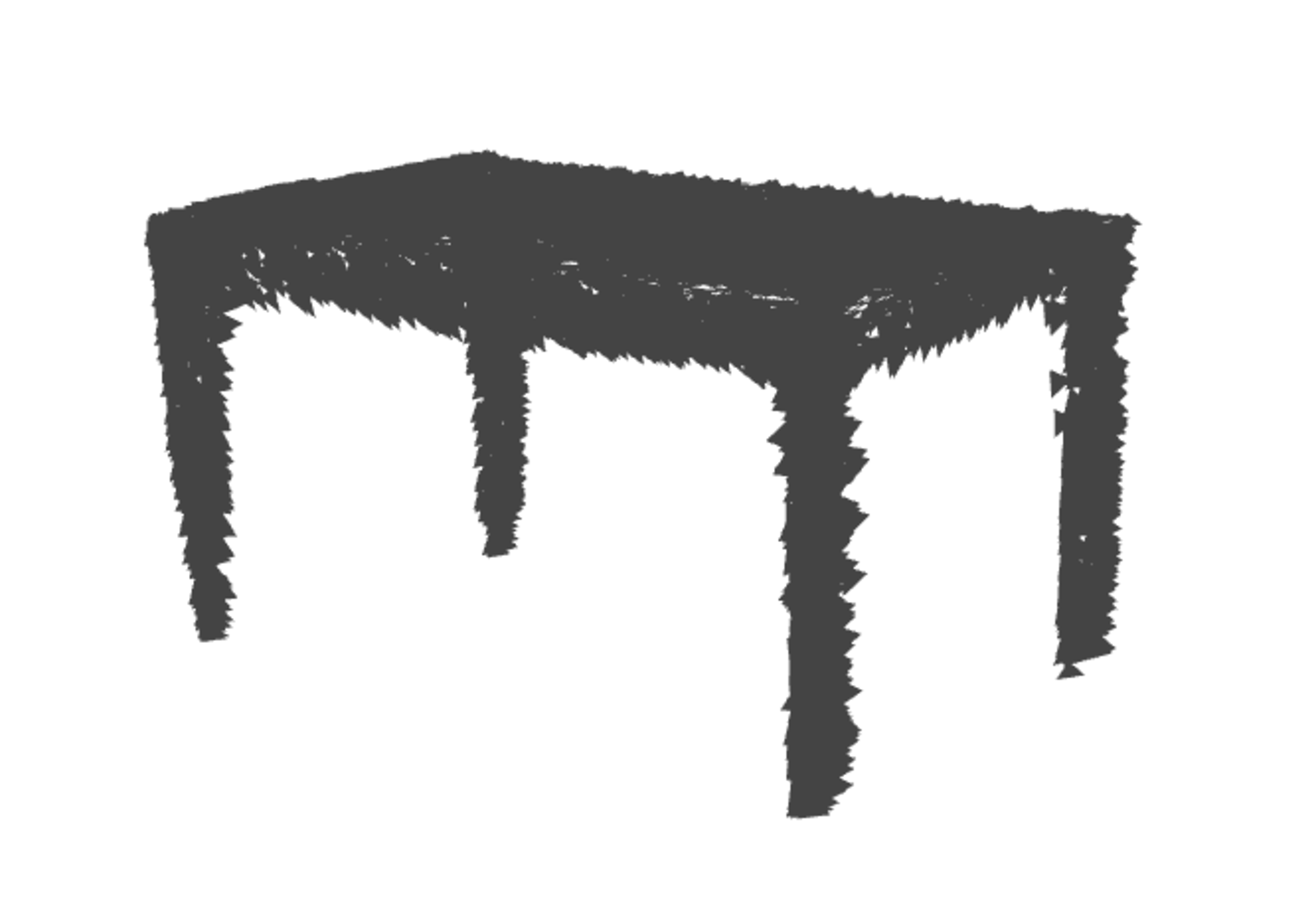} & 
  \includegraphics[width=0.23\textwidth]{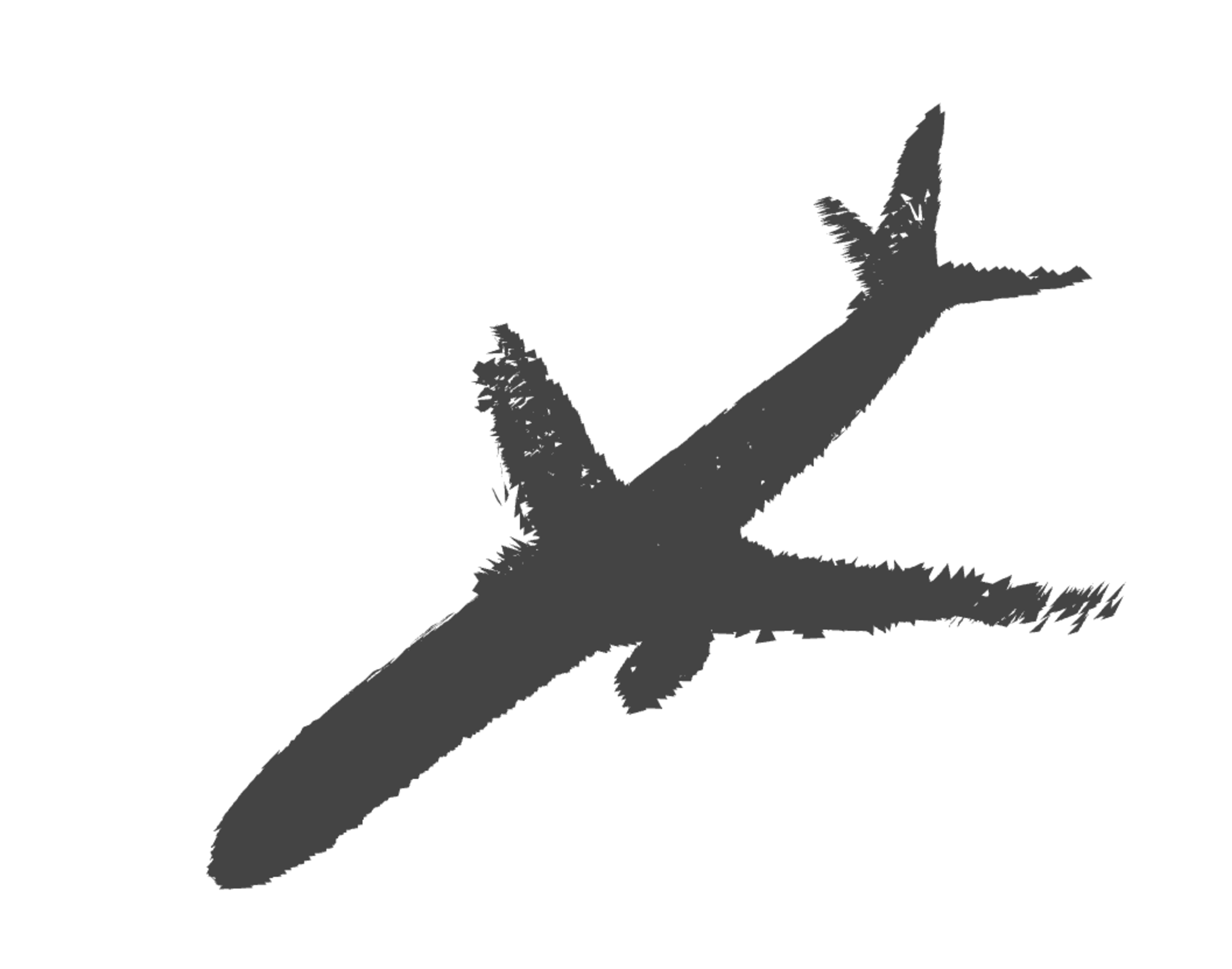} 
  \\\\\\

\end{tabular}
  \caption{Surface representation from known objects from four categories: chairs, lamps, tables and planes from left to right. Each tangent plane is drawn as a separate triangle patch. \label{tab:qual_triangle}}
\end{table*}

\subsection{Dataset}
For evaluation, we use the publicly available ShapeNet v1 dataset \cite{chang2015shapenet}, following the train/test splits provided by 3D-R2N2 \cite{choy20163d}. Specifically, we use the same rendering as Pixel2Mesh\cite{wang2018pixel2mesh}. The object is placed in a view-centric coordinate where it is rotated based on the rendering camera angle. A face normal vector of  a triangle is used as the ground truth surface normal vectors of every point sampled from the triangle.

\subsection{Metrics} 
\paragraph{Chamfer distance} We use Chamfer distance to assess the positional error between the generated surface points and the ground truth surface. We sample 2500 points from our input spherical domain and map the points onto the surface of the object. We also uniformly sample 10000 points from the object using Trimesh~\cite{trimesh}. The Chamfer distance is calculated between the predicted point set and the ground truth point set. Asymmetric Chamfer distance is defined as:

\begin{equation}
CD(X, Y) = \frac{1}{|X|}\sum_{x\in X}{\min_{y\in Y}{||x - y||_2^2}} \label{eq:1}
\end{equation}

$X$ and $Y$ are two unordered sets of points. In Table~\ref{tab:meancompare} and \ref{tab:chamfer_p2m}, we report the average of the prediction-to-ground-truth and ground-truth-to-prediction distances. We notice some works use an alternative formulation of Chamfer distance where absolute value of distance is used rather than the squared distance. Here, we follow the convention of \cite{wang2018pixel2mesh}. All numbers are multiplied by 1000 for ease of comparison.

\paragraph{F-1 Score} 
In the context of shape reconstruction, F-1 scores are defined using a distance threshold. A predicted point is considered correctly classified if there is a ground truth point within the threshold. This classification between ground truth and predicted point clouds is analogous to precision and recall. F-1 score is the harmonic mean of the two distances. Table~\ref{tab:f1} reports F-1 score comparisons with Pixel2Mesh. Since the F-1 scores reported in Mesh R-CNN are obtained using a rescaled dataset, we are not able provide comparisons with them at this point.

\paragraph{Surface Cosine Similarity} To evaluate local surface reconstruction, we measure the cosine similarity between the surface normal of the predicted local tangent plane and the ground truth surface normal. Note that for each surface at a single point, there are two valid surface normal directions that define the same local tangent plane. Following \cite{park2019deepsdf}, we calculate the cosine similarity between the predicted normal and the two possible normal directions and take the absolute value. For each vector, the cosine similarity is in range $[0, 1]$. Higher cosine similarity indicates better alignment of the predicted tangent plane with the ground truth surface. We uniformly sample 2500 points on the ground truth surface along with their surface normal vectors. For each sampled point, we search for the closest point in the predicted point cloud based on the Euclidean distance. In Table~\ref{tab:meancomparecos} and \ref{tab:tabcos_p2m} we compare the mean cosine similarity of the whole point set. The surface cosine similarity between two sets is defined as:
\begin{align}
c.sim(X_{gt}, X_{pred}) &= \frac{1}{|X_{gt}|}\sum_{x\in X_{gt}}{|{\Vec{n_x}\cdot{\Vec{n}_{\theta(x, X_{pred})}}}|} \\
\theta(x, Y) &= \argmin_{y \in Y} ||x - y||_2^2 
\end{align}
We compare our surface cosine similarity results with methods with two types of output: point set and mesh. For surface-based methods, we sample uniformly on the generated surface to get a point set. The surface normal of a triangle is used for all the points sampled from this triangle. Then we compute the set cosine similarity between the ground truth point set $X_{gt}$ and the point set sampled from the predicted mesh $X_{pred}$. On the other hand, to compare with point set based methods, we run the local surface normal estimation algorithm based on the Principle Components Analysis (PCA) of the local neighborhood. We use the implementation from \cite{Zhou2018} with default parameters.

\section{Results} In this section, we show the direct comparison of our method with several competing methods.

\begin{table}
\centering
\begin{tabular}{c c c c}
\hline
 \thead{P2M Code}  & \thead{Mesh \\ R-CNN\cite{gkioxari2019mesh}} &  \thead{\ourmethod{}} \\
\hline
 0.736 &  0.729 & \textbf{0.757}\\
\hline
\end{tabular}
\caption{Mean cosine surface distance comparison.\label{tab:meancomparecos}}
\end{table}

\begin{table*} 
\begin{tabular}{c c c c}
\includegraphics[width=0.23\textwidth]{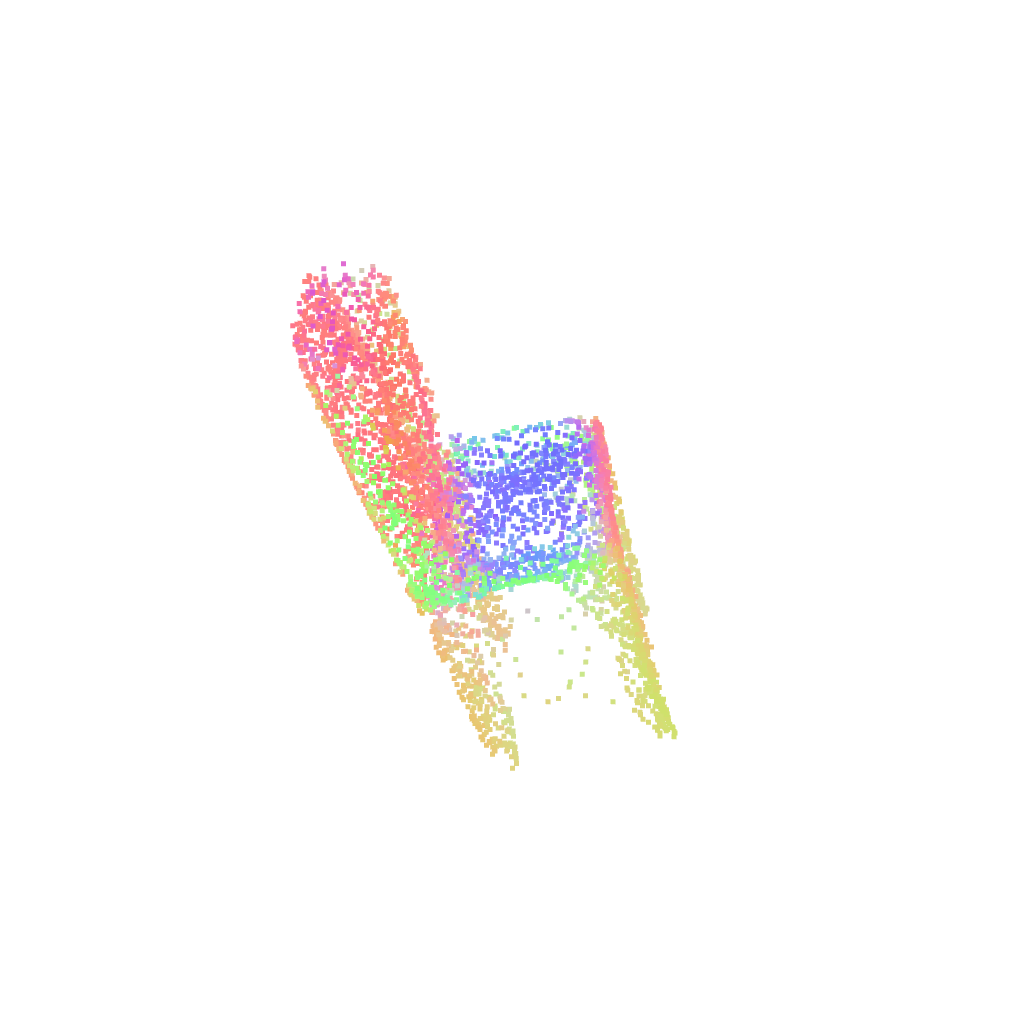} & \includegraphics[width=0.23\textwidth]{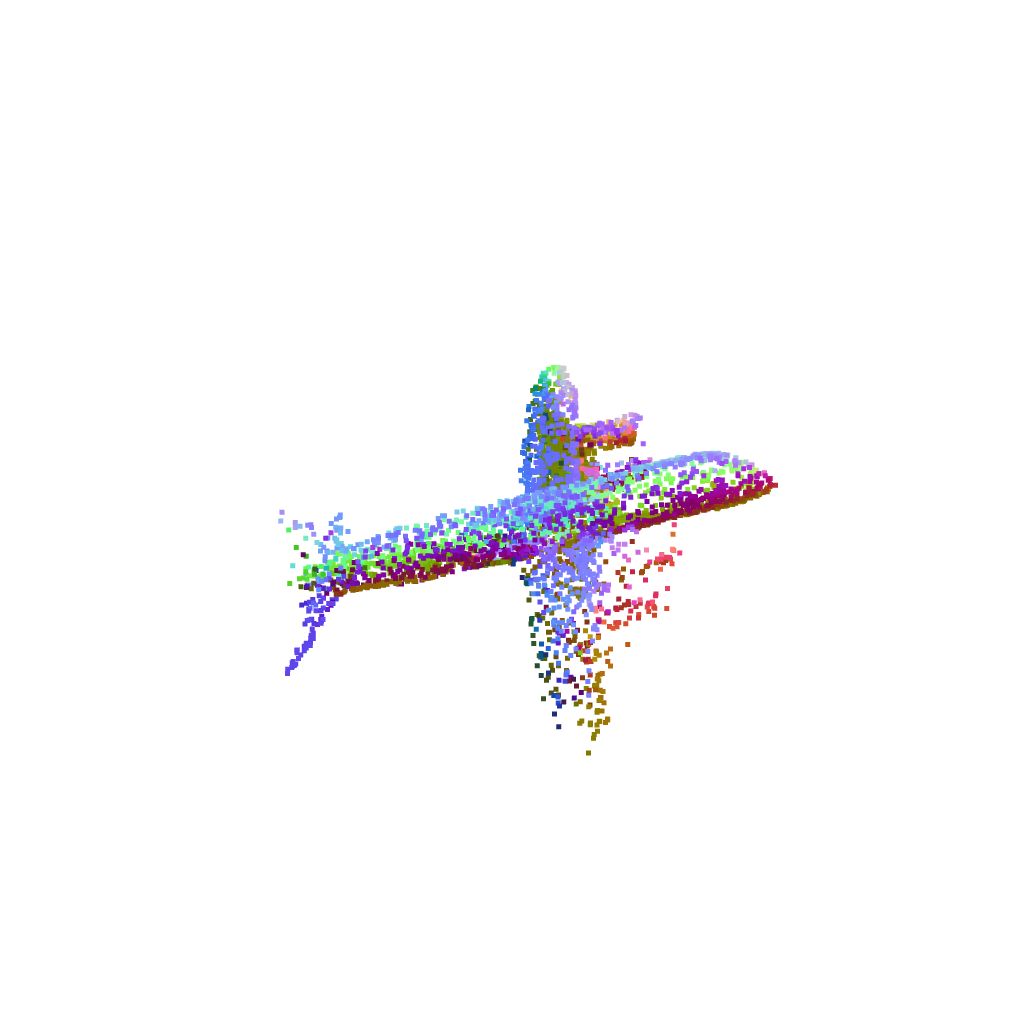} & \includegraphics[width=0.23\textwidth]{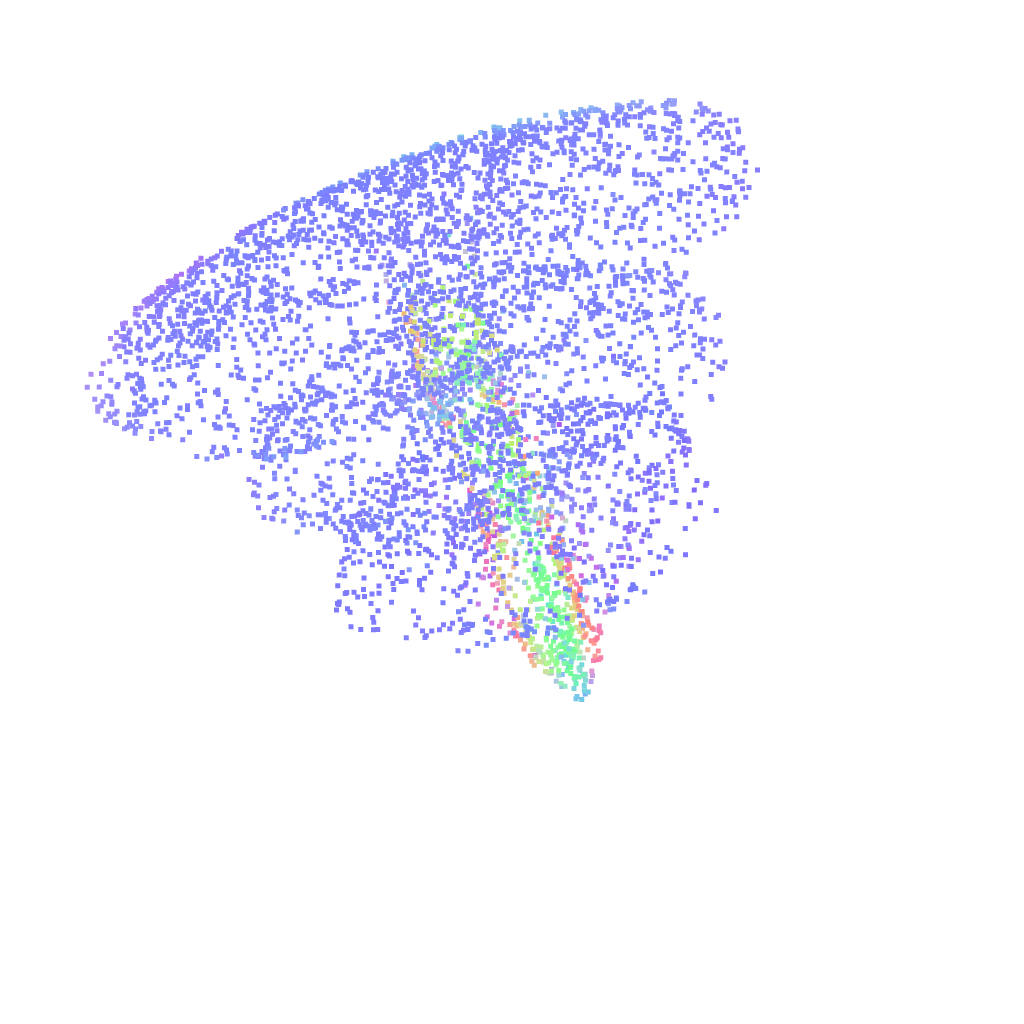} & \includegraphics[width=0.23\textwidth]{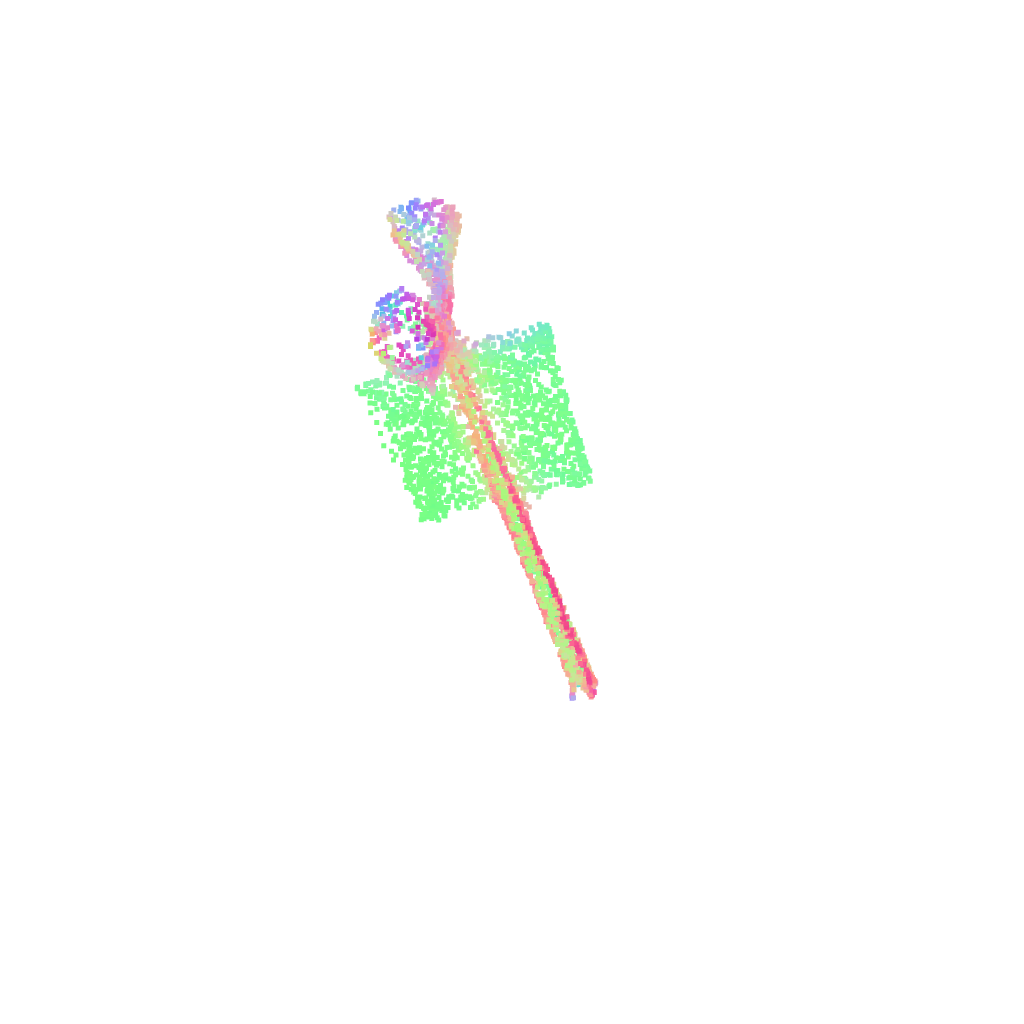} \\ 
  
  \includegraphics[width=0.23\textwidth]{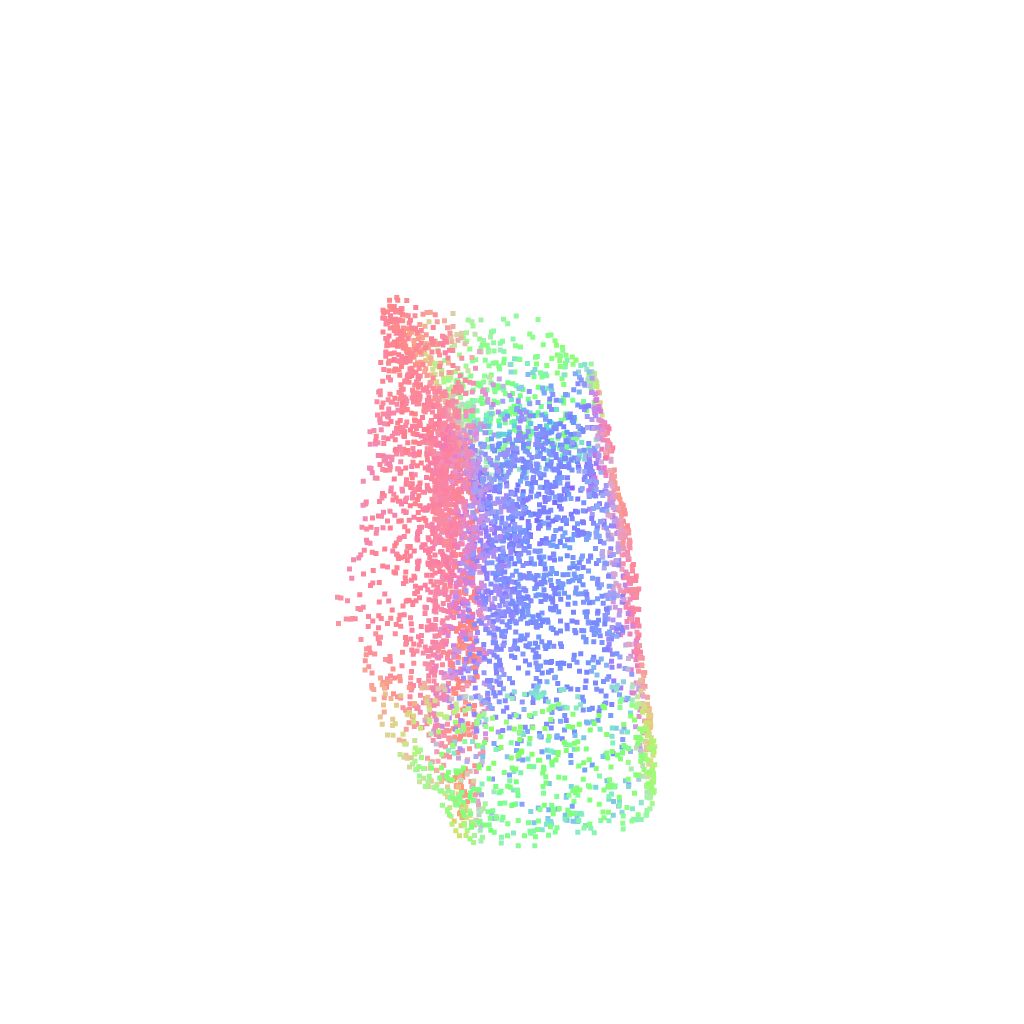} & \includegraphics[width=0.23\textwidth]{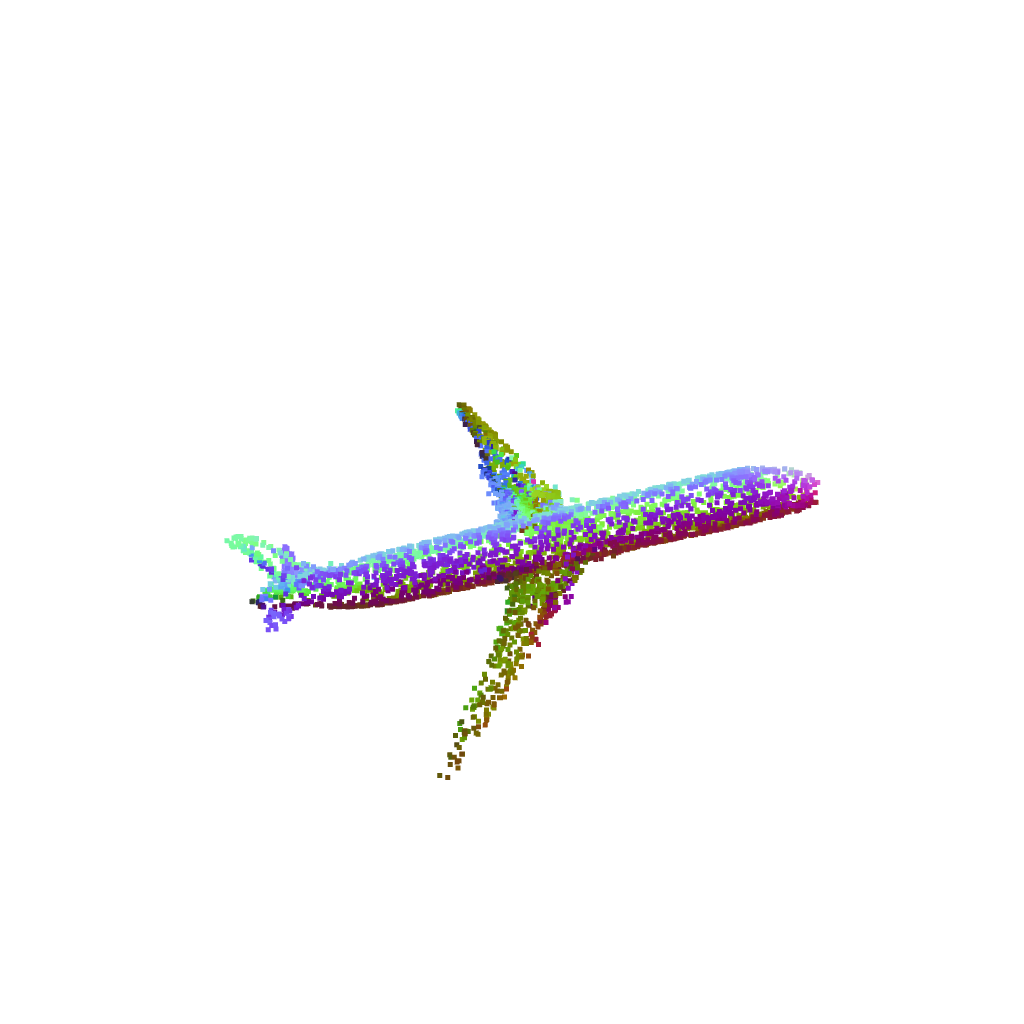} &  \includegraphics[width=0.23\textwidth]{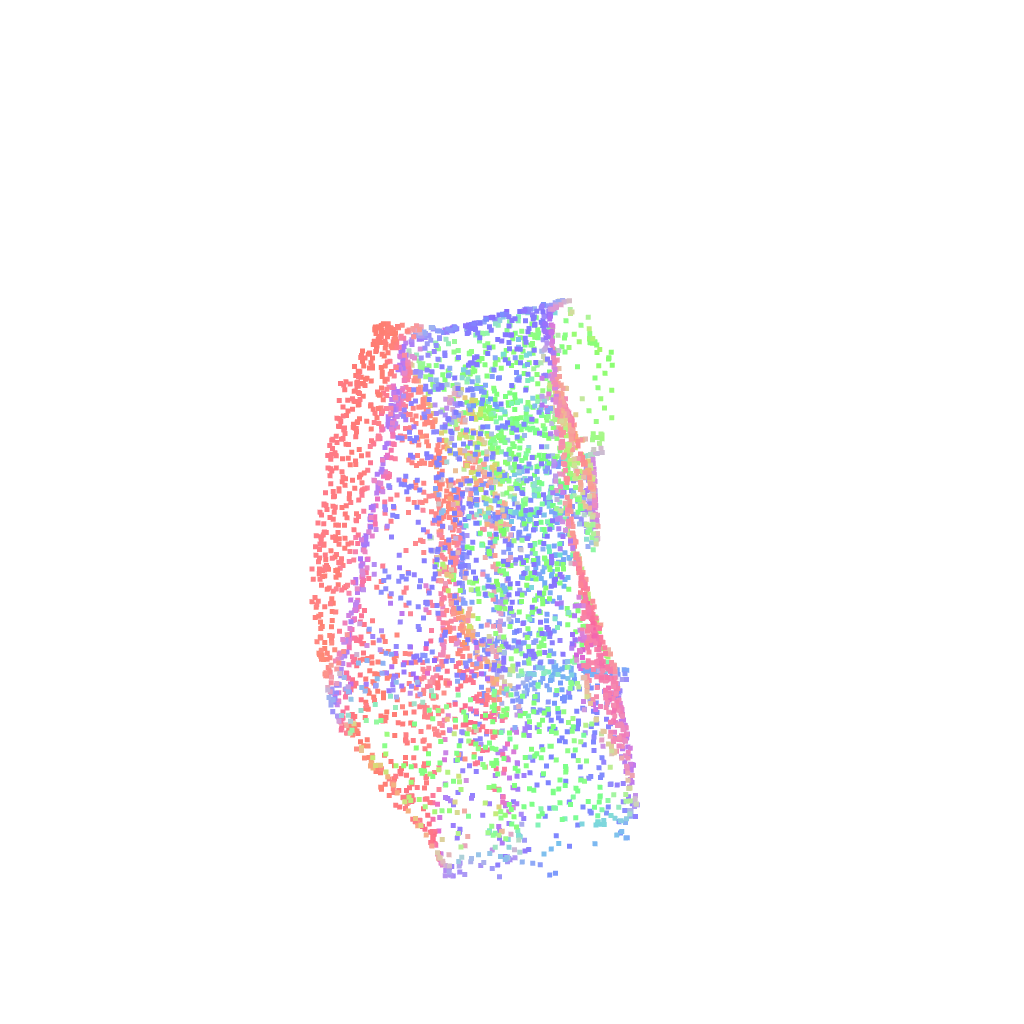} & \includegraphics[width=0.23\textwidth]{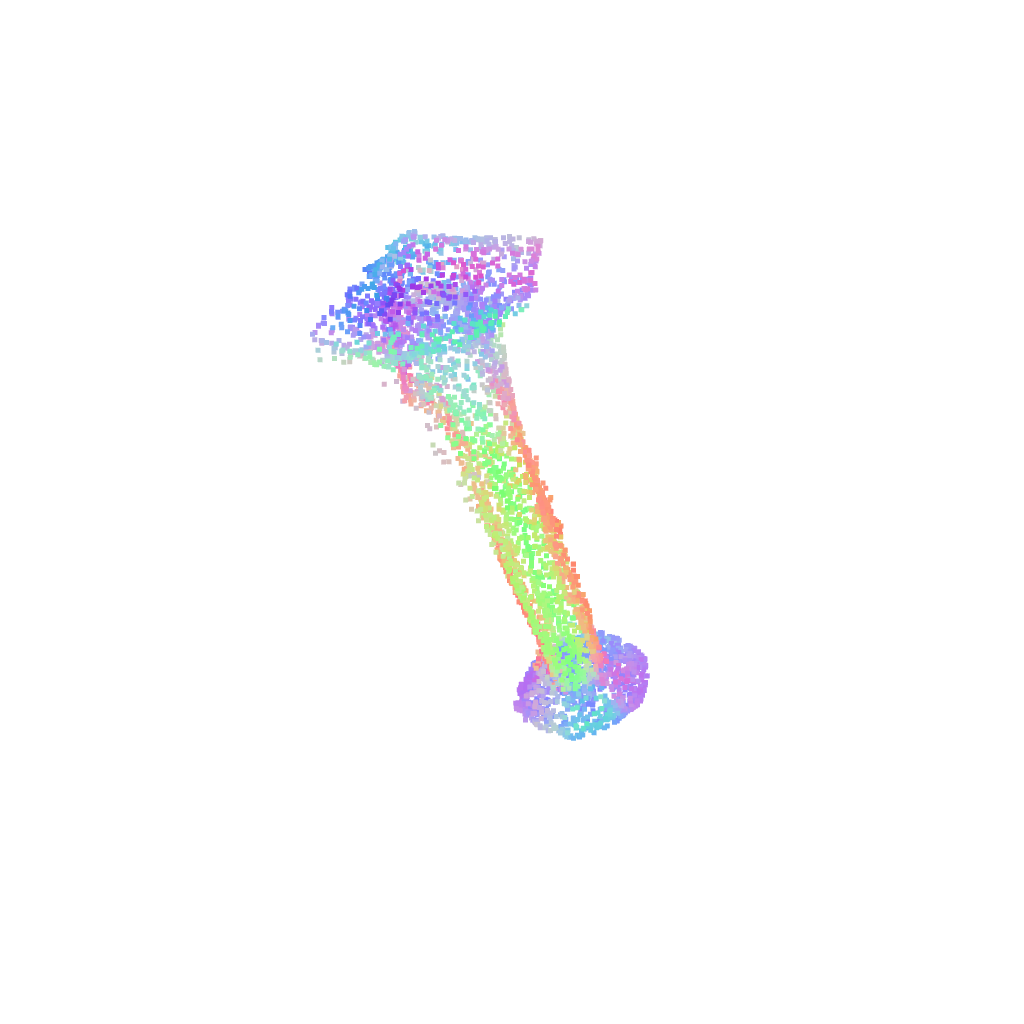} \\ 
  
  \includegraphics[width=0.23\textwidth]{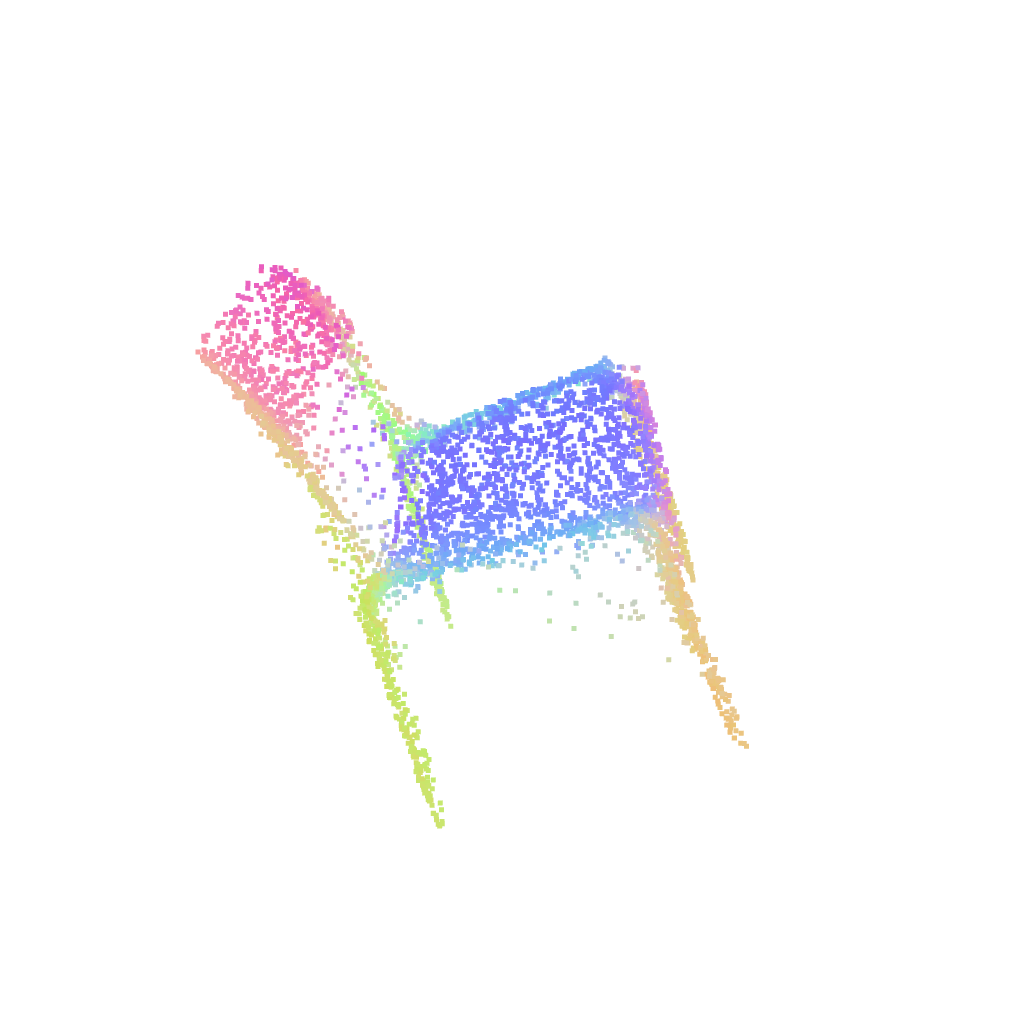} & \includegraphics[width=0.23\textwidth]{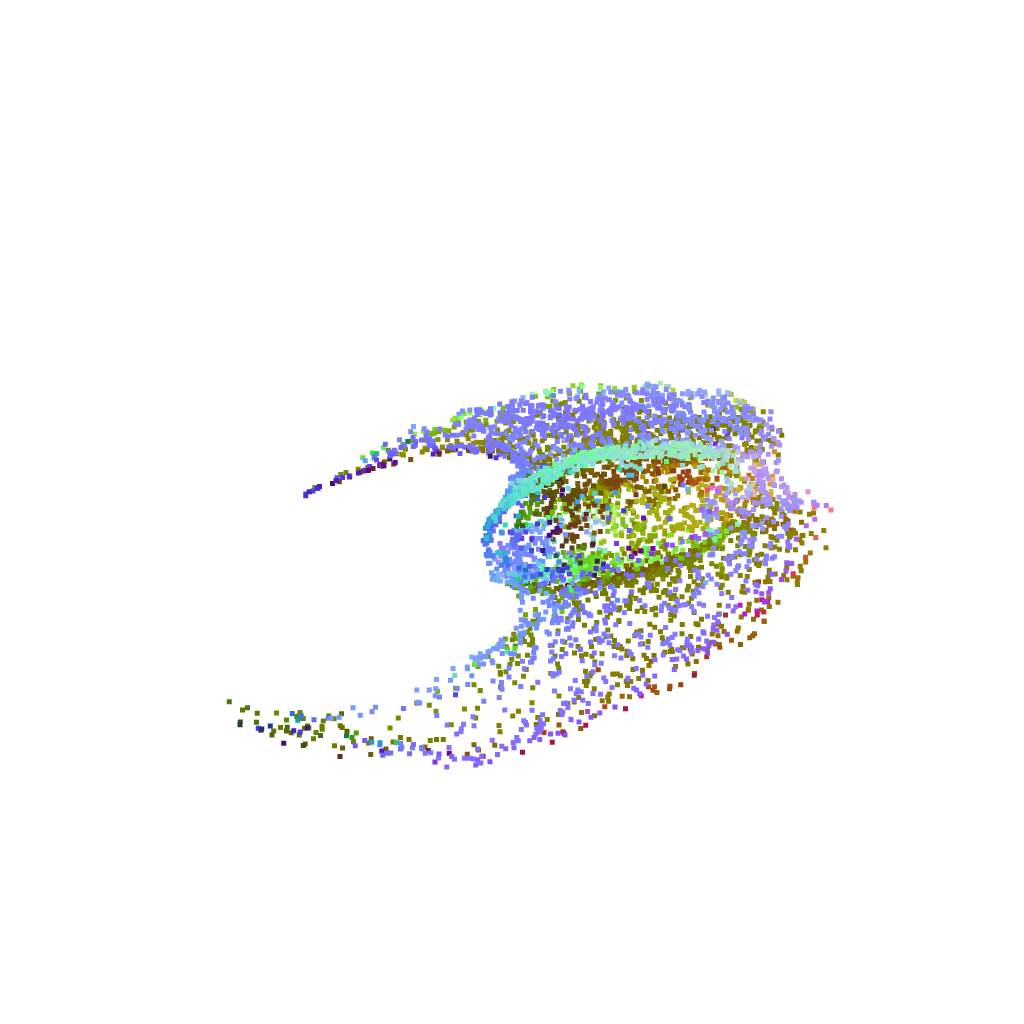} & \includegraphics[width=0.23\textwidth]{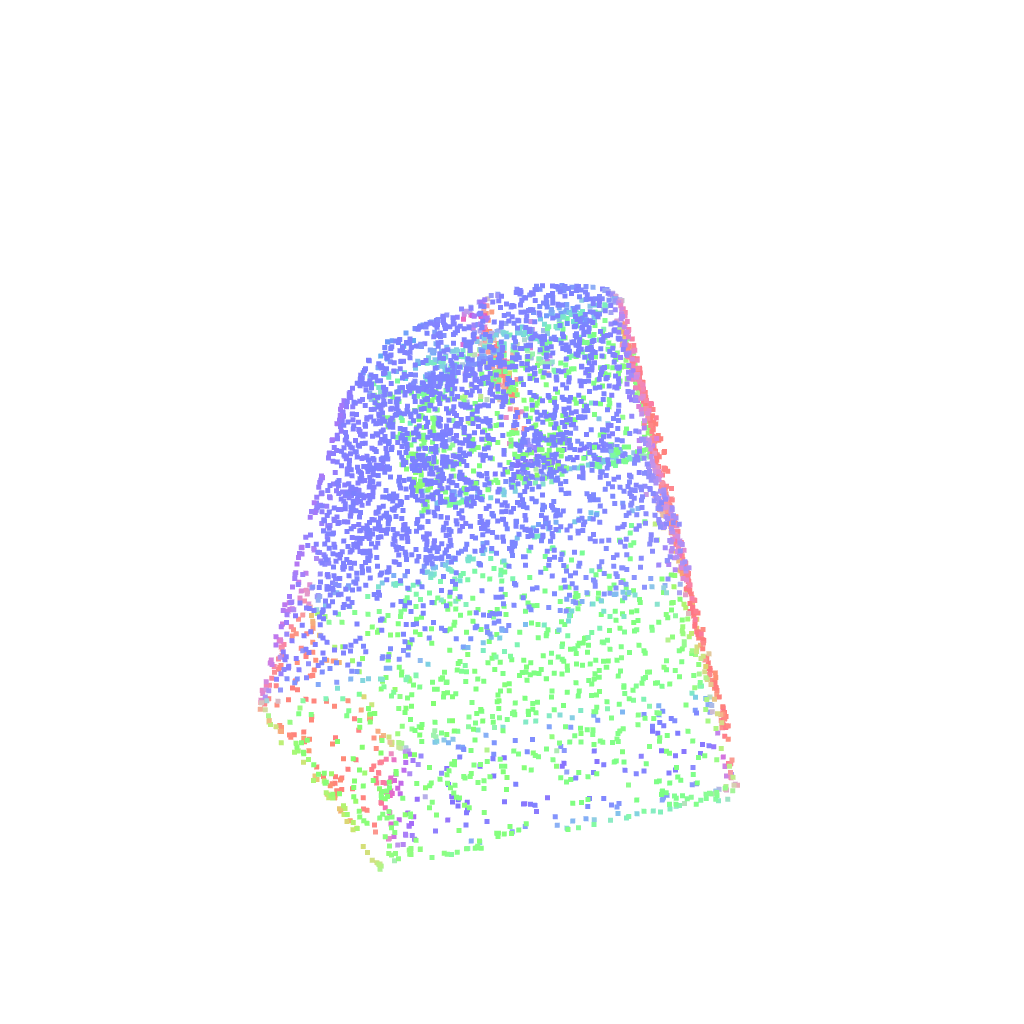} & \includegraphics[width=0.23\textwidth]{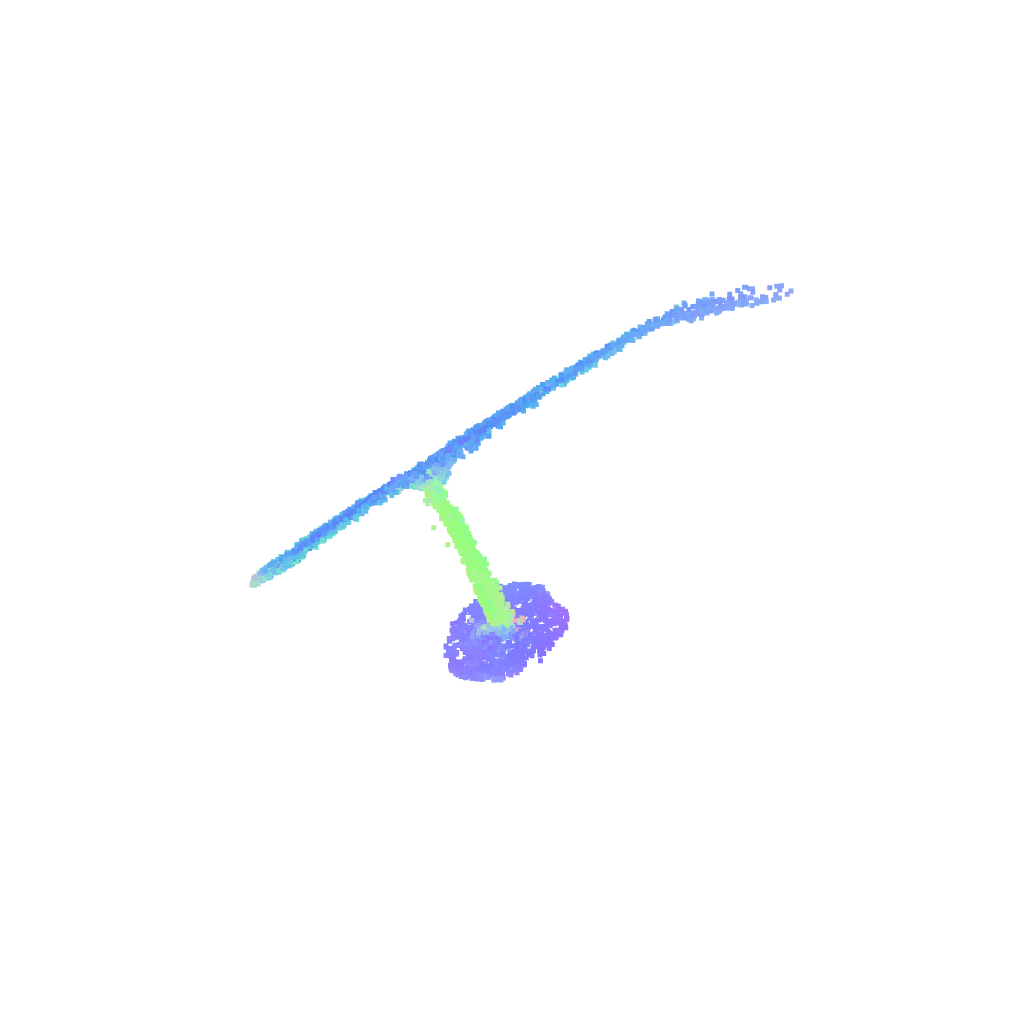} \\
  
  \includegraphics[width=0.23\textwidth]{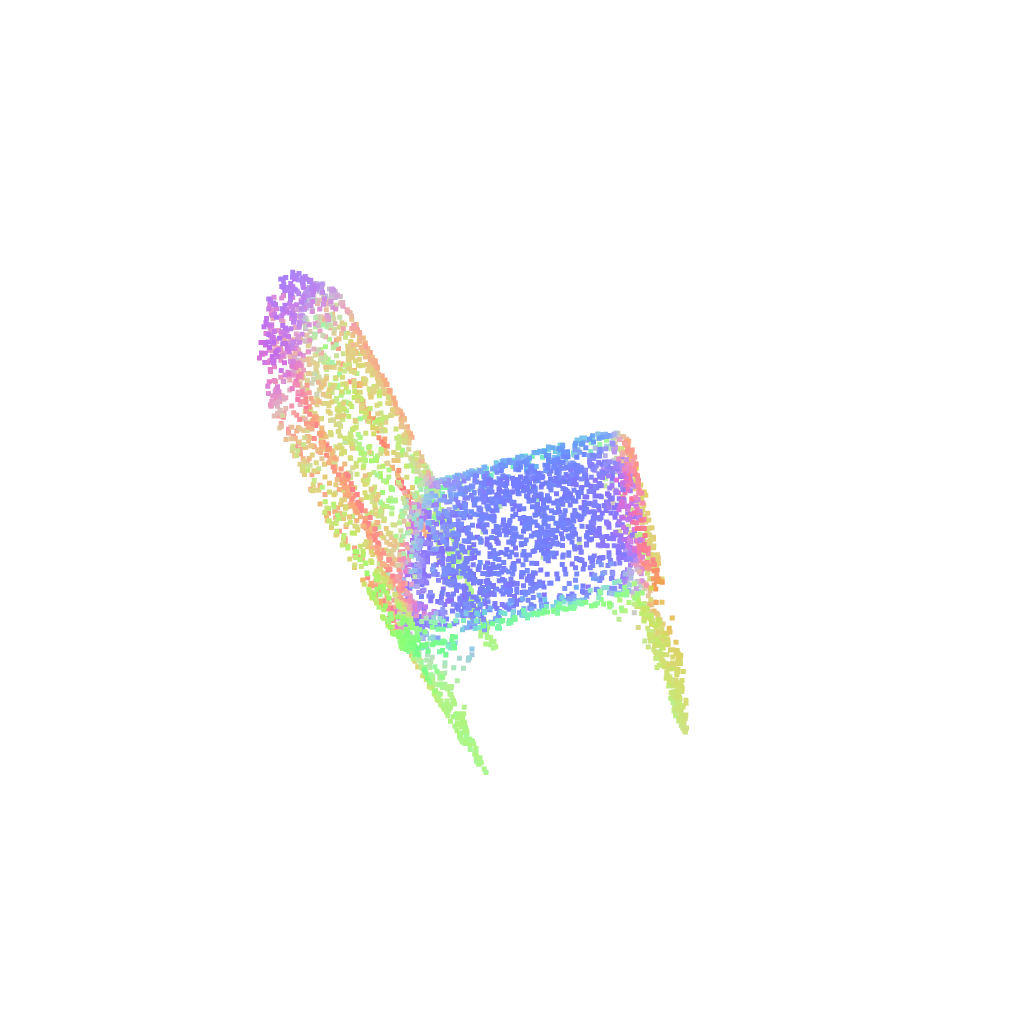} & \includegraphics[width=0.23\textwidth]{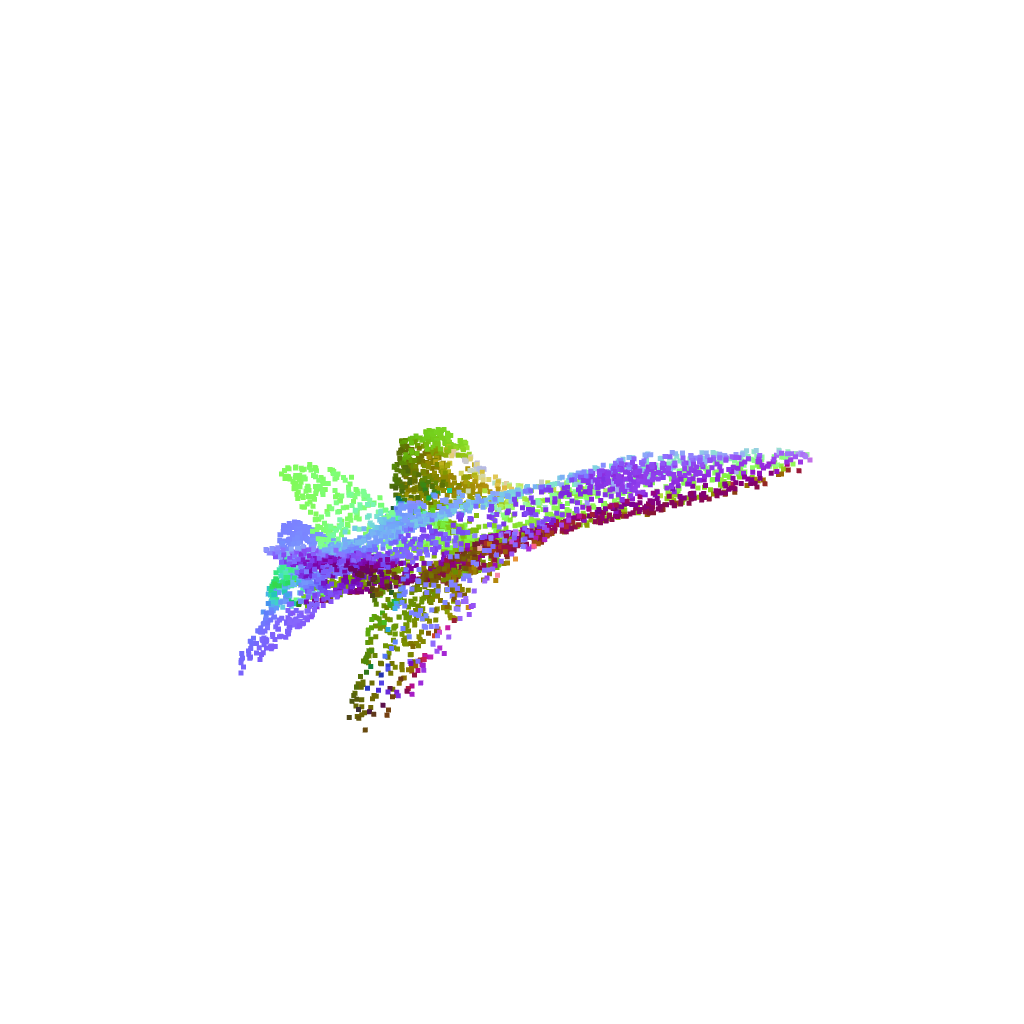} & \includegraphics[width=0.23\textwidth]{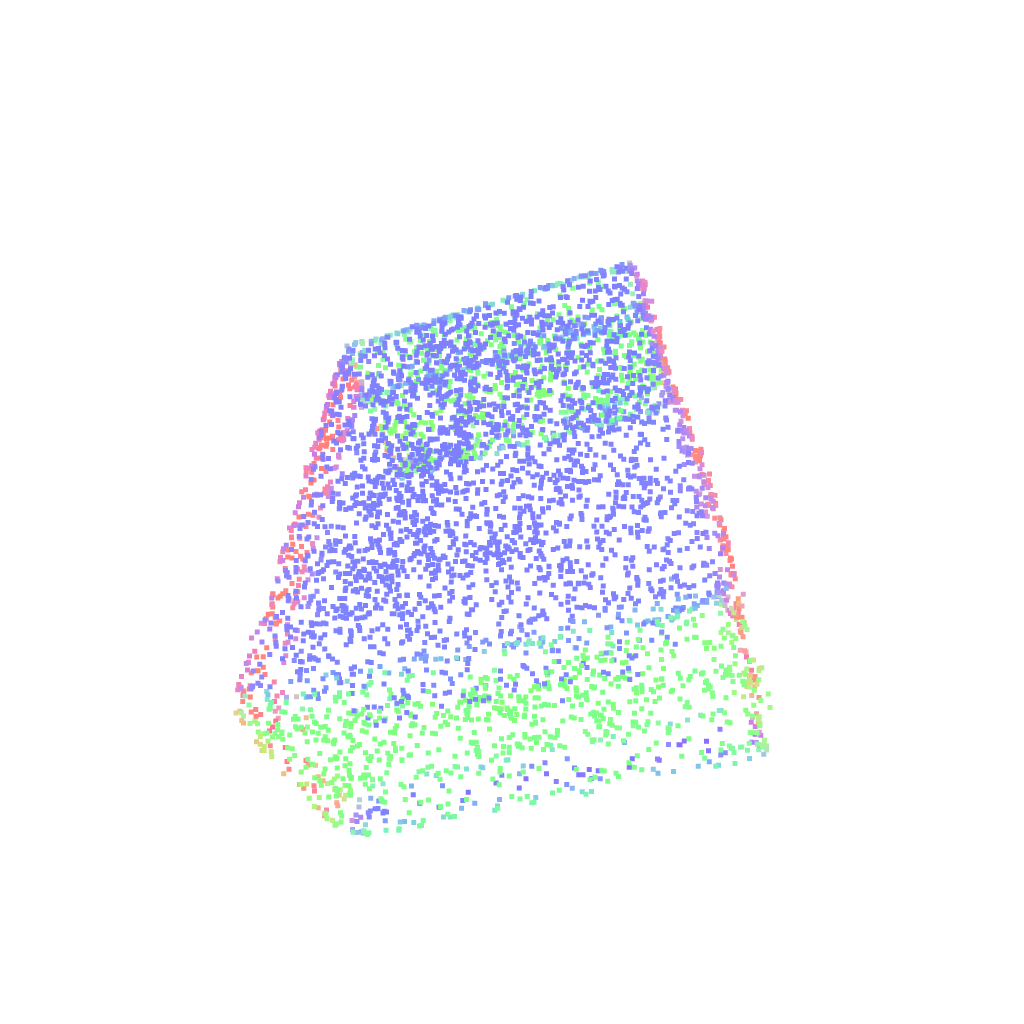} & \includegraphics[width=0.23\textwidth]{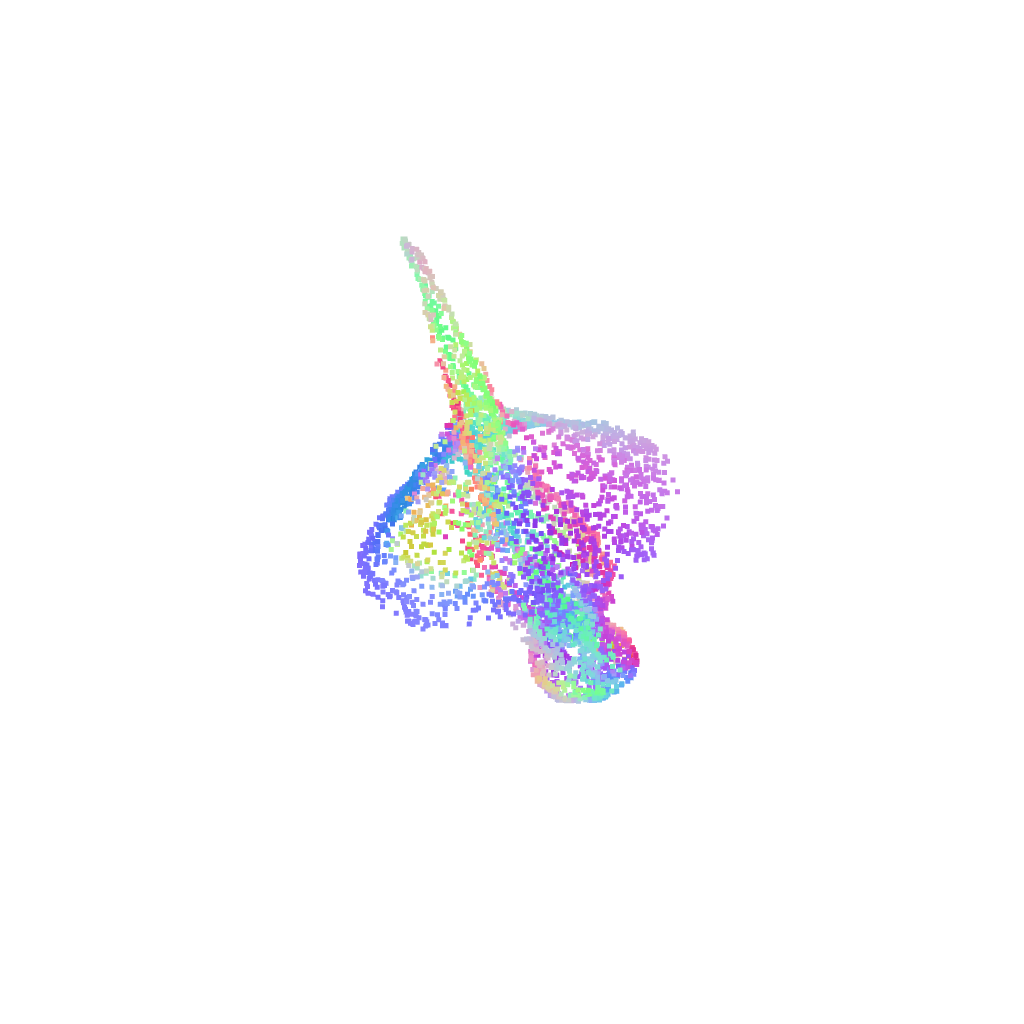} \\ 
  
  \includegraphics[width=0.23\textwidth]{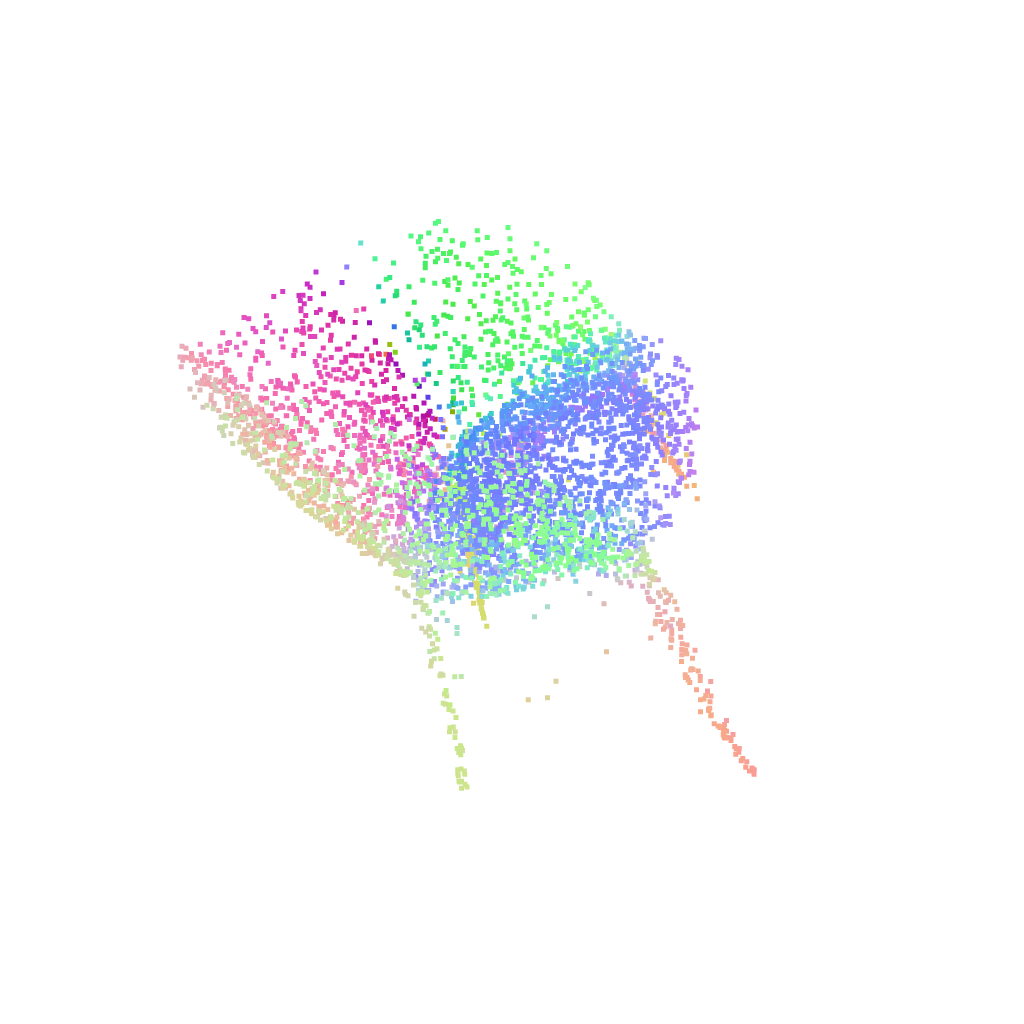} & \includegraphics[width=0.23\textwidth]{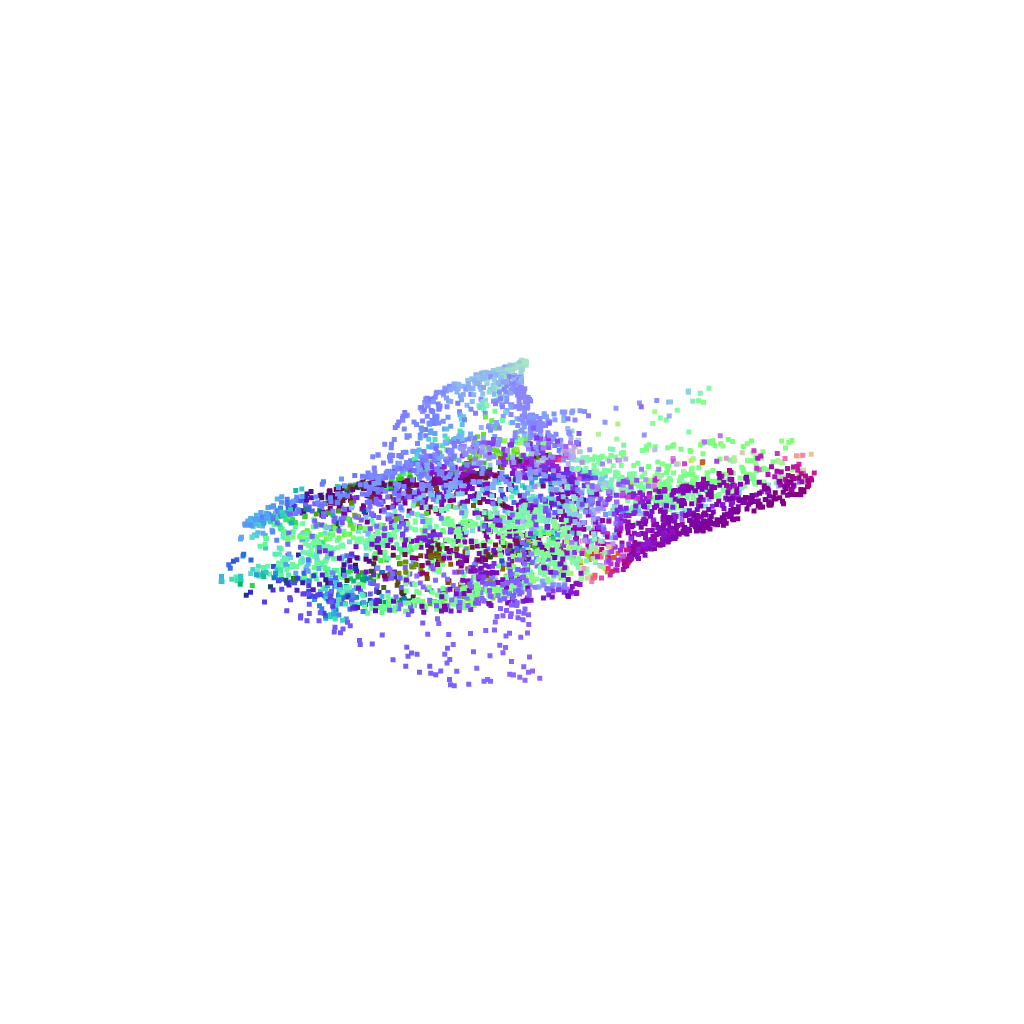} & \includegraphics[width=0.23\textwidth]{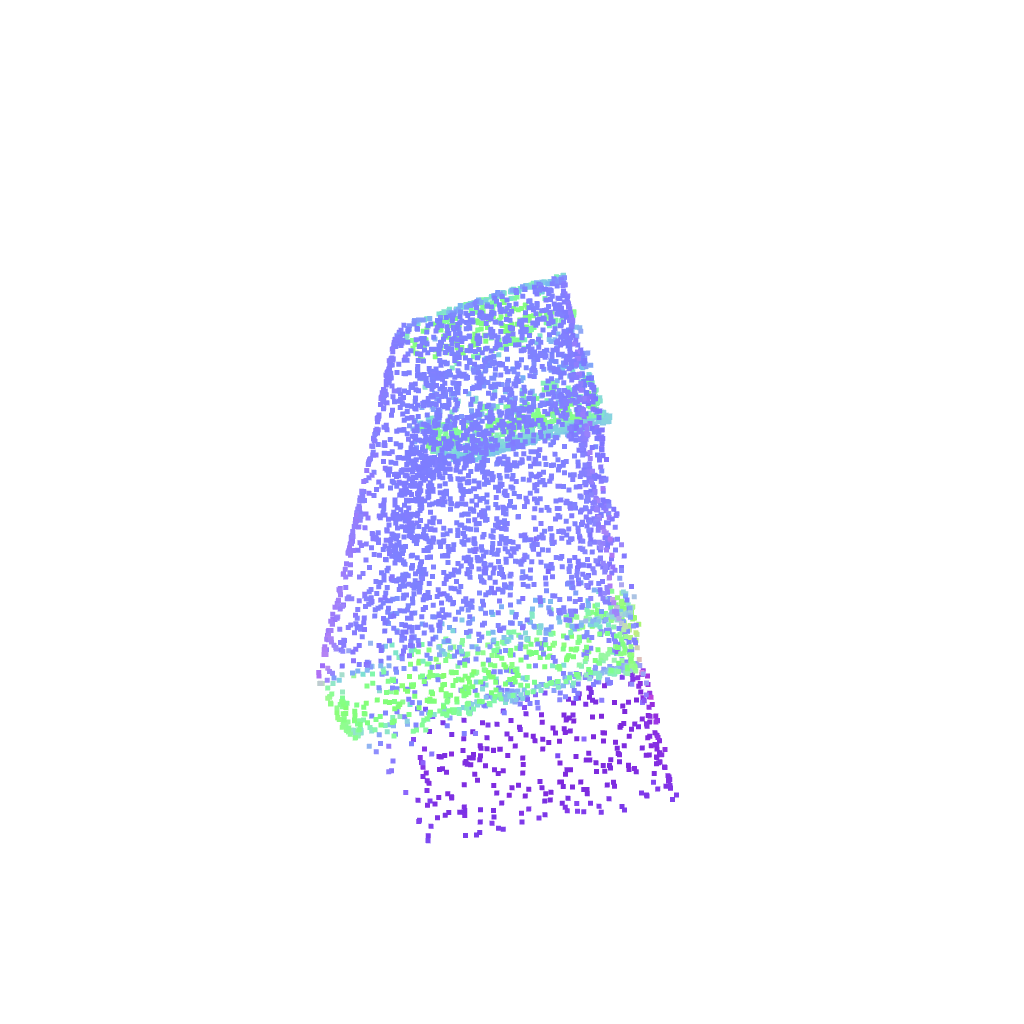} & \includegraphics[width=0.23\textwidth]{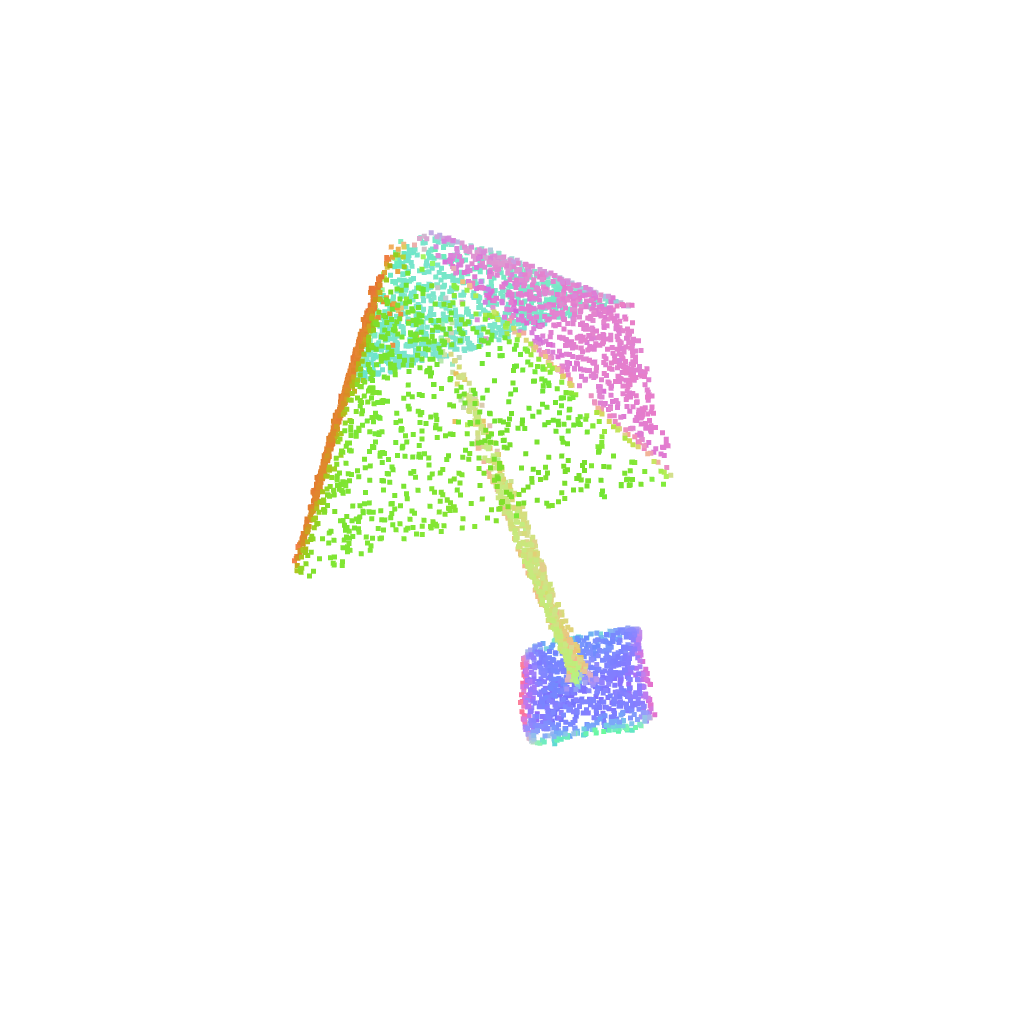} \\ 
\end{tabular}
  \caption{Reconstruction of known objects from four categories: chairs, airplanes, tables and lamps from left to right. Color represents the predicted surface normal direction. \label{tab: qualitative}}
\end{table*}

\paragraph{Point Evaluation} 
We compare our method with Pixel2Mesh (P2M)\cite{wang2018pixel2mesh}, Neural 3D Renderer (N3MR)\cite{kato2018neural}, PSG\cite{fan2017point} and Mesh R-CNN\cite{gkioxari2019mesh} in terms of Chamfer distance.  \\

The network is trained on the train split provided by 3D-R2N2\cite{choy20163d} and the dataset provided by Pixel2Mesh\cite{wang2018pixel2mesh}. We ran the publicly available evaluation code in Pixel2Mesh's online repository with the provided pre-trained weights. The test results are better than their reported numbers in \cite{wang2018pixel2mesh}. Therefore we report their numbers reported in the paper as P2M and the output of their pre-trained model as P2M Code. \\

Table~\ref{tab:chamfer_p2m} shows our method outperforms all methods in all categories by a significant margin in Chamfer distance. All the numbers are multiplied by 1000 for ease of comparison. This shows that \ourmethod{} is able to reconstruct the shape of the object effectively. In Table~\ref{tab:f1}, we report the per-category and total mean F-score under two different thresholds: $\tau_1=10^{-4}$ and $\tau_2=2\times10^{-4}$. From the results, our method performs better in all categories for a higher threshold. For the lower threshold, the results are mixed. Pixel2Mesh performs better in capturing the geometry of simpler objects while our method performs better on more detailed models.

\paragraph{Surface Evaluation} While Chamfer distance provides a measure of the similarity of two point sets, it does not fully capture the quality of surface reconstruction. Characteristics such as enclosure and collision are not well-represented in a sparse point cloud. Local tangent planes, however, provides a convenient way to encode the local surface geometry around a point on the object. Therefore, we compare the surface normal estimation quality of Surface HOF with two state of the art methods: Pixel2Mesh~\cite{wang2018pixel2mesh} and Mesh R-CNN~\cite{gkioxari2019mesh}.

\begin{table}
\centering
\begin{tabular}{l c c c c}
\hline
Category & P2M Code & Point HOF + PCA & \ourmethod{}  \\
\hline \hline
plane       & 0.768 & 0.754 & \textbf{0.804}\\
bench       & 0.676 & \textbf{0.751} & 0.713\\
cabinet     & 0.788 & 0.690 & \textbf{0.823}\\
car         & 0.710 & \textbf{0.746} & 0.731\\
chair       & 0.713 & 0.736 & \textbf{0.751}\\
monitor     & 0.824 & 0.714 & \textbf{0.822}\\
lamp        & 0.662 & \textbf{0.708} & \textbf{0.708}\\
speaker     & \textbf{0.788} & 0.715 & \textbf{0.788}\\
firearm     & 0.643 & \textbf{0.714} & 0.669\\
couch       & 0.781 & 0.738 & \textbf{0.782}\\
table       & 0.746 & 0.763 & \textbf{0.835}\\
cellphone   & \textbf{0.889} & 0.709 & 0.856\\
watercraft  & 0.695 & \textbf{0.720} & 0.705\\
\hline
mean  & 0.745 & 0.728 & \textbf{0.771}\\
\hline
\hline
\end{tabular}
\caption{Cosine similarity result evaluated on the ShapeNet Test set. Reconstructed from single RGB image. Trained and tested with Pixel2Mesh\cite{wang2018pixel2mesh} splits. Higher is better.
\label{tab:tabcos_p2m}
}
\end{table}

To compare against Pixels2Mesh, we ran their evaluation code, sampled points from their generated mesh and extracted the surface normal of each point from the triangle it belongs to. For fair comparison, we uniformly sampled 2500 points from the mesh surface generated by Pixel2Mesh and compute the cosine surface distance. 

In Table~ \ref{tab:meancomparecos}, we compare mean cosine surface similarity with Mesh R-CNN as reported by the authors in their paper. Despite its simple architecture, \ourmethod{} outperforms both methods.

Per class results are reported in Table~\ref{tab:tabcos_p2m}. It can be seen that our surface normal outperforms Pixel2Mesh in almost all categories other than cellphone. To demonstrate that Surface HOF captures surface properties better than our previous work~\cite{anon2019} which generates only point clouds, we also report results from  a version of our architecture (Point HOF) which only outputs a point. Point HOF is trained using only the Chamfer Distance. Once the point cloud is estimated, normals are computed using local PCA. The results are reported in the column Point HOF PCA. It can be observed that, while the average performance of this method is lower, there are certain categories where the surface normal estimation from point cloud is better than both Pixel2Mesh and \ourmethod{}. This further shows that we can locally estimate good surface information given the ability to generate the surface point set at arbitrary resolution. 

\section{Qualitative Results}

In Figures~\ref{tab: qualitative} and ~\ref{tab:qual_triangle}, we show qualitative results of our reconstruction of various types of objects in ShapeNet. Each point defines a local tangent plane that defines the local geometry of the surface. In Figures~\ref{tab: qualitative}, color represents the orientations of the local tangent plane. In Figure~\ref{tab:qual_triangle}, tangent planes are drawn as triangles without triangulation. Generally, we should expect two things from a good reconstruction by looking at this visualization: (i)~The points are consistently mapped to the surface without too many outliers, and (2)~the points on the same surface should have equal or similar color, indicating the local consistency of surface normal orientation.


\section{Concluding Remarks}

In this paper, we presented Surface HOF as an approach for generating high quality surfaces. For a given input image, Surface HOF generates a mapping function which can be used for generating detailed surfaces at arbitrary resolution. 
The method is efficiently trained using supervised backpropagation by optimizing global shape and normal information while obeying local geometrical constraints. 
It is not limited to specific topologies such as genus zero surfaces and, as shown by the experiments, can handle a number of complex geometries. It achieves better or comparable performance to other state of the art methods while using significantly less number of parameters. 

The HOF approach is conceptually related to  the `fast-weight' paradigm~\cite{schmiudhuber1992learning} and  `hypernetworks'~\cite{ha2016hyper}. Similar to our paper, these earlier works also advocate the advantages of functional representations in terms of parameter efficiency. Additional applications illustrating the advantages of the HOF approach are shown in the Supplementary materials. Theoretical investigation into justifying the overall efficiency and performance of the method is an important avenue for future research. On the more practical side, we are interested in extending the HOF paradigm to output additional object information such as colors and textures, and using it for closed-loop robotic tasks such as navigation and object manipulation.

{\small
\bibliographystyle{ieee_fullname}
\bibliography{refs}
}

\end{document}